\documentclass[lettersize,journal]{IEEEtran}
\usepackage{amsmath,amssymb,amsfonts}
\usepackage{algorithm}
\usepackage{array}
\usepackage[caption=false,font=normalsize,labelfont=sf,textfont=sf]{subfig}
\usepackage{textcomp}
\usepackage{stfloats}
\usepackage{url}
\usepackage{verbatim}
\usepackage{graphicx}
\usepackage{cite}

\usepackage{xcolor}
\usepackage{tabularx}
\usepackage{multirow}
\usepackage{algpseudocode}  
\usepackage{adjustbox}
\usepackage{makecell}
\usepackage{booktabs} 
\usepackage{wrapfig}        
\usepackage{listings} 

\definecolor{LightGray}{gray}{0.9}

\begin{document}

\title{DIAL: Distribution-Informed Adaptive Learning of Multi-Task Constraints for Safety-Critical Systems}

\author{Se-Wook Yoo$^{1}$, ~\IEEEmembership{Member,~IEEE}, and Seung-Woo Seo$^{1}$, ~\IEEEmembership{Member,~IEEE}
    \thanks{Manuscript received December 6, 2024.}
    \thanks{This work was supported by the BK21 FOUR program of the Education and Research Program for Future ICT Pioneers, Seoul National University, and
    the Challengeable Future Defense Technology Research and Development Program through the Agency For Defense Development (ADD) funded by the Defense Acquisition Program Administration (DAPA) in 2024 (No.915108201).}
     \thanks{$^{1}$The authors are with the Department of Electrical and Computer Engineering and ASRI, Seoul National University, Seoul, Republic of Korea.
        {\tt\footnotesize \{tpdnr1360, sseo\}@snu.ac.kr}}
    }



\maketitle

\begin{abstract}
\label{chapter3:abstract} Safe reinforcement learning has traditionally relied on predefined constraint functions to ensure safety in complex real-world tasks, such as autonomous driving. However, defining these functions accurately for varied tasks is a persistent challenge. Recent research highlights the potential of leveraging pre-acquired task-agnostic knowledge to enhance both safety and sample efficiency in related tasks. Building on this insight, we propose a novel method to learn shared constraint distributions across multiple tasks. Our approach identifies the shared constraints through imitation learning and then adapts to new tasks by adjusting risk levels within these learned distributions. This adaptability addresses variations in risk sensitivity stemming from expert-specific biases, ensuring consistent adherence to general safety principles even with imperfect demonstrations. Our method can be applied to control and navigation domains, including multi-task and meta-task scenarios, accommodating constraints such as maintaining safe distances or adhering to speed limits. Experimental results validate the efficacy of our approach, demonstrating superior safety performance and success rates compared to baselines, all without requiring task-specific constraint definitions. These findings underscore the versatility and practicality of our method across a wide range of real-world tasks.
\end{abstract}

\begin{IEEEkeywords}
Deep Learning in Robotics and Automation, Learning from Demonstration, Robot Safety, Robotics in Hazardous Fields.
\end{IEEEkeywords}

\section{introduction}
\label{chapter3:introduction}

\IEEEPARstart{A}{cquiring} comprehensive knowledge \cite{laskin2021cic, laskin2021urlb, mutti2022unsupervised} has been a central focus in the fields of deep reinforcement learning (RL) \cite{sutton2018reinforcement} and imitation learning (IL) \cite{pomerleau1991efficient, schaal1996learning, argall2009survey}, as it enables solving complex real-world problems. Recent advances in task-agnostic exploration \cite{finn2017model, badia2019never, tao2020novelty, seo2021state} demonstrate that leveraging shared knowledge across diverse tasks can significantly enhance performance in downstream tasks. However, in safety-critical domains such as autonomous driving, unrestricted exploration is infeasible \cite{garcia2015comprehensive, marot2020learning, dulac2021challenges}. Safe RL \cite{achiam2017constrained, wachi2020safe, qin2021density} addresses this by defining safely explorable regions as constraints. Learning a safe exploration policy \cite{yang2023cem} not only ensures adherence to safety requirements but also promotes effective transferability to novel tasks, enabling the agent to explore states within the boundaries of predefined constraints. Nevertheless, crafting accurate constraint functions across diverse tasks presents a burden. Moreover, relying solely on the acquired policy without constraints during transfer learning (TL) risks losing essential safety information previously learned.

The inverse constraint RL (ICRL) framework \cite{chou2020learning, malik2021inverse} offers a solution by alternatively learning constraints to ensure safety while simultaneously developing the policy that achieves goals safely under the defined reward function. However, when these constraints are restored solely from single-task demonstrations, unexplored areas in the policy space are considered unsafe. This can lead to discrepancies between inferred and actual constraints, resulting in conservative policies with large regret bounds. The issue is further exacerbated in downstream tasks, where overly conservative constraints impede finding feasible solutions. To reduce regret bounds, incorporating multi-task demonstrations to learn constraints is preferable \cite{kim2024learning}. Nonetheless, collecting demonstrations across diverse tasks poses practical challenges because it requires meeting multiple safety requirements and addressing biases that arise from experts' risk tendencies, which remain significant hurdles in existing ICRL methods.

\begin{figure*}[t]
    \centering        
    \subfloat[Standard ICRL]{        
        \centering
        \begin{minipage}{0.94\textwidth}             
            \includegraphics[width=0.4\textwidth]{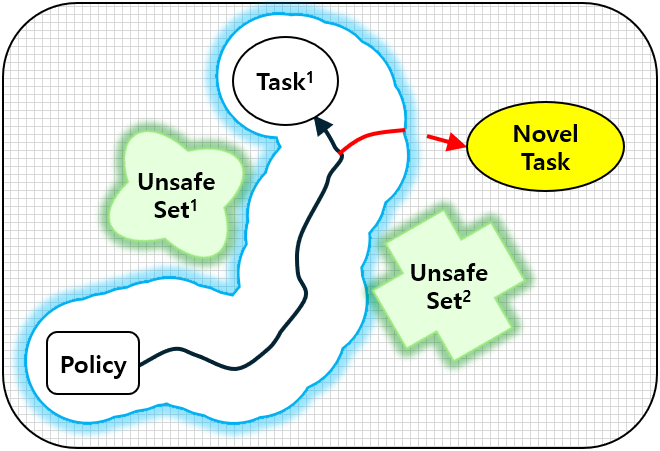} 
            \label{fig1a:stadard_icrl_problem}
            \hfill
            \includegraphics[width=0.5\textwidth]{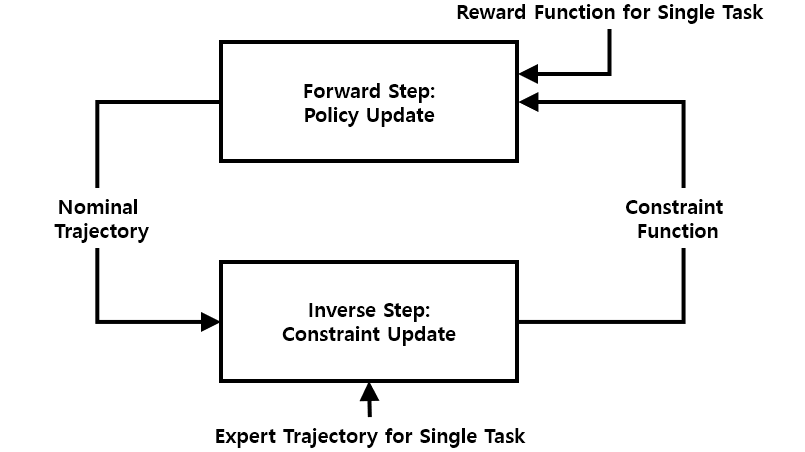} 
            \label{fig1a:stadard_icrl_architecture}
        \end{minipage}        
    }\hfill
    \subfloat[Proposed ICRL]{        
        \centering
        \begin{minipage}{0.94\textwidth}
            \includegraphics[width=0.4\textwidth]{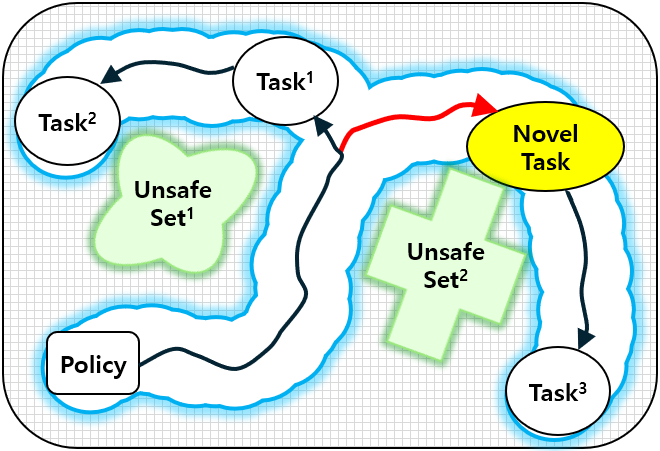} 
            \label{fig1b:proposed_icrl_problem}
            \hfill
            \includegraphics[width=0.5\textwidth]{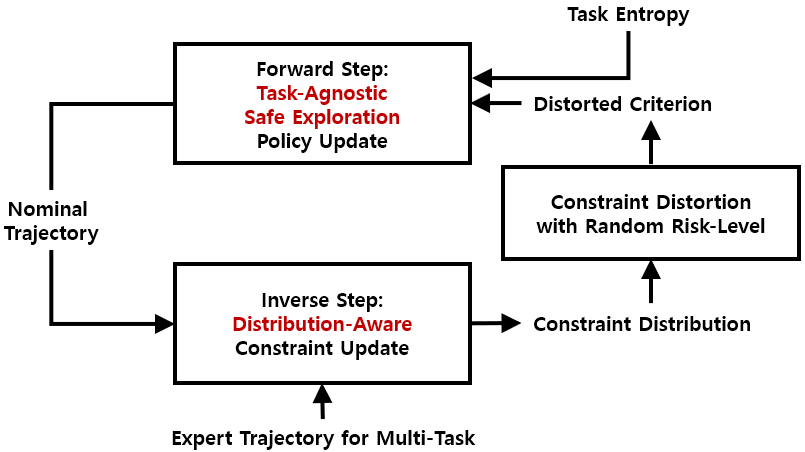} 
            \label{fig1b:proposed_icrl_architecture}
        \end{minipage}
    }    
    \caption{Comparison of standard ICRL (a) and proposed ICRL with DIAL (b). The left side shows the problem each approach addresses. In (a), the black arrow represents the expert policy avoiding unsafe sets (green polygons) and staying within the attraction region (blue boundary), but standard ICRL cannot adapt to novel tasks as shown by the red arrow. In (b), DIAL learns a distribution-aware constraint function from multi-task demonstrations to adapt safety constraints for new environments and uses task-agnostic safe exploration to enable safe adaptation across tasks. These improvements help DIAL find feasible solutions for novel tasks, as highlighted by the red arrow, with red-coded components on the right illustrating its architectural differences over standard ICRL.}
    \label{fig1:overview_of_proposed_ICRL}
\end{figure*}

To overcome these limitations, we propose distribution-informed adaptive learning (DIAL), a novel method that leverages shared knowledge across diverse tasks to enable safe and effective adaptation to novel tasks. DIAL extends the ICRL framework by incorporating a distributional understanding of risks inherent in multi-task demonstrations. This risk-awareness is implemented through distorted criteria such as conditional value at risk (CVaR), allowing dynamic adjustment of risk levels to facilitate safe adaptation to changing environments \cite{khokhlov2016conditional, pmlr-v100-tang20a, yang2021wcsac}. The core idea of DIAL is to design the constraint function to capture the distribution of risk across tasks while encouraging task-agnostic safe exploration (TASE) by maximizing entropy across diverse risk levels. Fig. \ref{fig1:overview_of_proposed_ICRL} describes an overview of DIAL. Similar to standard ICRL, DIAL alternatively learns both the constraint function and the policy from demonstrations. However, DIAL introduces two critical innovations: 1) In the inverse step, DIAL learns the constraint distribution from the multi-task demonstrations, providing rich supervision of safety requirements. 2) In the forward step, DIAL maximizes task entropy within the learned risk bounds, encouraging safe exploration across a broader range of tasks. These innovations bring several benefits. The learned constraint distribution facilitates risk adjustment through distorted criteria, enabling adaptation to changed safety conditions, as illustrated by the green polygons on the left side of Fig. \ref{fig1b:proposed_icrl_architecture}. TASE policy also enables agents to effectively find feasible solutions for TL to meta-task scenarios, as depicted by the red arrow.


Our work offers three main contributions:

\begin{itemize}
\label{contribution}

\item We propose DIAL by incorporating awareness of constraint distributions from multi-task demonstrations and TASE into ICRL. This enables scalable knowledge acquisition that ensures compliance with safety requirements and supports solving multiple tasks.

\item We also alleviate the burden of manually designing cost functions by capturing constraint distributions through restored functions that align with actual safety constraints across various tasks.

\item Safe adaptation to novel tasks in shifted safety conditions is achieved by applying risk-sensitive constraints to guide TASE policy exploration while correcting biases in demonstrations. This enables DIAL to excel in safety-critical TL benchmarks, showcasing strong real-world potential.

\end{itemize}

\section{Related Work}
\label{related_work}

Building on recent advancements in RL and IL, this research is inspired by various methodologies that enhance task adaptability and safety in autonomous systems.


\textbf{Learning Safe Exploration Policy:} Task-agnostic exploration \cite{zhang2020task, lindner2024learning, greenberg2024train} has become essential in RL to build generalized knowledge across diverse tasks without reliance on specific rewards or environmental dynamics. Early efforts \cite{laskin2021cic, laskin2021urlb, mutti2022unsupervised} established frameworks for learning exploration policy that adapts effectively to new tasks. Most techniques in this area focus on improving exploration efficiency through deep RL or IL \cite{tao2020novelty,stadie2015incentivizing, jin2020reward}. While these methods excel in adaptability, they typically do not address the safety constraints critical to real-world applications. On the contrary, our study aims to address through a safety-focused exploration strategy. Safe exploration policy balances the need for broad state exploration with constraints that maintain system safety. Previous works \cite{hazan2019provably, lee2019efficient} have proposed state density maximization as a means to enable exploration but often limited its application to discrete state spaces. Some studies \cite{wen2018constrained, yang2023cem} on maximizing constrained entropy illustrate how safety constraints can be incorporated into RL frameworks to ensure TASE while enabling TL across different domains. Our study builds upon these ideas by proposing a model that uses IL rather than RL to learn safe exploration policy, reducing the need for precise cost functions and improving sample efficiency.
\setlength{\parskip}{0pt}


\textbf{Learning Reward Function:} Traditional IL methods \cite{sammut1992learning, hayes1994robot, abbeel2004apprenticeship}, which align policy with expert behaviors through supervised learning, often encounter compounding errors, especially in dynamic settings. Advanced hybrid methods \cite{ho2016generative, fu2018learning, garg2021iq} based on maximum entropy IRL (MaxEnt IRL) \cite{ ziebart2008maximum} have mitigated these issues by unifying IL and RL approaches \cite{piot2016bridging, finn2016connection}. However, adversarial IRL approaches such as GAIL \cite{ho2016generative} and AIRL \cite{fu2018learning} present stability and interpretability issues in reward function learning. To address these issues, assuming that the reward function is known, we modify the ICRL framework \cite{malik2021inverse} that avoids adversarial training and focuses on interpretable constraint learning that can generalize across safety-critical tasks.


\textbf{Learning Constraint Function:} The shift from reward learning to constraint learning arises from the need to ensure safety rather than merely replicating expert behaviors. Approaches \cite{scobee2019maximum, malik2021inverse} that use maximum likelihood inference, a variant of IRL, aim to learn constraint conditions from expert demonstrations without predefined functions. However, these methods can be sensitive to data quality and may struggle to generalize constraint conditions across multiple tasks. Other approaches \cite{chou2020learning, chou2021learning, chou2022gaussian} use grid-based or probabilistic parameters to define safety boundaries, guiding robots to avoid dangerous areas through model-based exploration. These methods show potential for maintaining constraint conditions across various tasks and enabling real-world robotic applications. However, high-resolution grid representations come with increased computational costs, which limit scalability. To address this, efforts have focused on developing flexible models capable of applying constraint functions to new environments. For instance, some studies \cite{gaurav2023learning, xu2023uncertainty, subramanian2024confidence} incorporate uncertainty into constraint inference from a distributional perspective, allowing learned constraints to adapt reliably to changed dynamics. Meanwhile, a reward-decomposition approach \cite{jang2023inverse} facilitates the safe transfer of constraints to new reward structures. While these studies improve the generalizability of learned constraint functions, they often remain limited to specific tasks. Prior works \cite{kim2024learning, lindner2024learning} address this limitation by exploring multi-task learning, but they still fail to account for distributional shifts. In contrast, our proposed approach learns an adaptive policy and constraint function that adjusts risk to ensure safety without strict assumptions tied to any specific task, providing a more flexible and broadly applicable solution.

Previous studies highlight the need for constraint learning that can be flexibly applied across diverse tasks. Our proposed approach, DIAL, emphasizes the derivation of a risk-sensitive constraint function and an adaptive policy from multi-task demonstrations, allowing for safety without being bound to specific tasks. DIAL effectively adjusts risk bias and supports policy adaptation across various scenarios. This design ensures scalability and safety, positioning it as a promising solution for applications such as autonomous driving and other safety-critical fields.

\section{Preliminaries}
\label{preliminaries}


\subsection{Maximum Entropy Constrained Reinforcement Learning}

The CRL approach considers the environment as a constrained Markov decision process (CMDP) \cite{altman2021constrained} to learn an optimal policy that maximizes the discounted cumulative rewards while ensuring the agent adheres to a set of safety constraints. To maintain consistency in notation, we redefine the CMDP from a distribution perspective as a following tuple $(\mathcal{S}, \mathcal{A}, \mathcal{P}_{T}, \mathcal{R}, \mu, \gamma, \mathcal{C})$. Here $\mathcal{S} \in \mathbb{R}^{\vert \mathcal{S} \vert}$ and $\mathcal{A} \in \mathbb{R}^{\vert \mathcal{A} \vert}$ are state and action space, respectively, where $s \in \mathcal{S}, a  \in \mathcal{A}$ are the elements for each space. $\mathcal{P}_{T}(s' \vert s, a)$ indicates the state transition dynamics. $\mathcal{R}$ represents the set of all reward functions, where $r(s,a) \in \mathcal{R}: \mathcal{S} \times \mathcal{A} \rightarrow \mathbb{R}$ is a reward function. Let $\mu$ denote the initial state distribution, $\gamma \in (0,1)$ the discount factor, and $\mathcal{C} = \{ (\mathcal{P}_{C_i}, \epsilon_{i} )\}_{i=1}^{K}$  the set of $K$ constraints. For the $i$-th constraint, $\mathcal{P}_{C_i} (c \vert s, a)$ gives the probability of incurring cost $c$ in state $s$ and action $a$, and $\epsilon_{i} \geq 0$ is a budget, which is an upper bound on expected cumulative costs. A cost function $c_{i}(s,a) \in C_{i}: \mathcal{S} \times \mathcal{A} \rightarrow \{0,1\}$ is represented as an indicator function for unsafe conditions, where $C_{i}$ is the set of all cost functions. We denote a policy as $\pi(a\vert s) \in \Pi : \mathcal{S} \times \mathcal{A} \rightarrow [0,1]$, which maps states to probability distributions over actions, where $\Pi$ is the set of all policies. 
Let us refer to the $\gamma$-discounted cumulative sum over an infinite time horizon as the return. To be specific, we denote reward-return as $r(\tau) = \sum_{t=0}^{\infty} \gamma^{t}r(s_{t},a_{t})$ and cost-return as $c_{i}(\tau) = \sum_{t=0}^{\infty} \gamma^{t}c_{i}(s_{t},a_{t})$, respectively. Here $\tau= (s_{0},a_{0},s_{1},a_{1}, \dots)$ is a trajectory of state-action pair and a set of trajectories is $\mathcal{D}=\{\tau\}_{j=1}^{N}$. When $\pi(\tau) = \mu(s_0) \prod_{t=0}^{\infty} \pi(a_t \vert s_t) \mathcal{P}_{T}(s_{t+1} \vert s_t, a_t)$ is defined as the probability that policy yield trajectory with $s_{0} \sim \mu$, $a_{t} \sim \pi(\cdot \vert s_{t})$, and $s_{t+1} \sim \mathcal{P}_{T}(\cdot \vert s_{t}, a_{t})$ for $t \geq 0$, we can represent the expected reward-return as $\mathbb{E}_{\tau \sim \pi(\cdot)} [r(\tau)]$ and a constraint with expected cost-return as $\mathbb{E}_{\tau \sim \pi(\cdot)} [c_{i}(\tau)] \leq \epsilon_{i}$, respectively. Note that this type of constraint is typically referred to as "soft", meaning that it allows some trajectories to exceed the cost-return threshold $\epsilon_i$, as long as the expected cost-return stays within $\epsilon_i$. In contrast, a "hard" constraint demands that each trajectory individually remains within the cost bound of $\epsilon_i$. To enforce a hard constraint, one could set $\epsilon_i = 0$ to ensure all costs remain non-negative. In this context, the objective of maximum entropy (MaxEnt) CRL is to determine an optimal policy that maximizes the expected entropy-regularized reward-return while satisfying the given constraints. This is represented by the following formulation \cite{achiam2017constrained}:

\begin{equation}
    \begin{aligned}
    \pi^* &= \arg \max_{\pi} \mathbb{E}_{\pi} \left[r(\tau) \right] + \beta \mathcal{H}(\pi) \\
    &\text{subject to} \quad \mathbb{E}_{\pi} \left[ c_{i}(\tau) \right] \leq \epsilon_i \quad \forall i,
    \end{aligned}
    \label{eq1:CRL_problem}
\end{equation}

where augmented term $\mathcal{H}(\pi) = - \int_{\tau} \pi(\tau) \log \pi(\tau) d\tau$ with coefficient $\beta \rightarrow \infty$ promotes randomness in action selection, supporting a certain level of exploration to prevent getting stuck in local optima while satisfying constraints.

\subsection{Inverse Constraint Reinforcement Learning}

In practical applications, it is challenging to manually specify all constraints to obtain an optimal policy in CRL. Nevertheless, it may be feasible to obtain expert demonstrations that satisfy safety requirements. ICRL effectively recovers the constraints from expert trajectories, assuming that the reward is available separately. To formalize this, previous studies \cite{scobee2019maximum, malik2021inverse} that build on our starting point extend the MaxEnt model \cite{ziebart2008maximum}. This model represents the probability of a trajectory under the policy $\pi$ and is adapted to CMDP as follows:

\begin{equation}
    \begin{aligned}
    P_{\pi}(\tau) &= \frac{\exp(\frac{1}{\beta} r(\tau)) \mathbf{1}(c_i(\tau) \leq \epsilon_i \, \forall i)}{Z(c_i)} \\
    \text{where } Z(c_i) &= \int_{\tau} \exp(\frac{1}{\beta} r(\tau)) \mathbf{1}(c_i(\tau) \leq \epsilon_i \, \forall i) \, d\tau.
    \end{aligned}
    \label{eq2:MaxEnt_model}
\end{equation}

Here $\mathbf{1}(c_i(\tau) \leq \epsilon_i \, \forall i)$ serves as a feasibility indicator, representing whether a trajectory meets all constraints. Since checking this indicator becomes intractable when the state and action spaces are continuous, it is replaced by a binary classifier $\zeta(\tau)$, which approximates the indicator using a differentiable neural network with sigmoid activation. This leads to the following maximum likelihood problem \cite{malik2021inverse}:

\begin{equation}
    \begin{aligned}
        \zeta^*(\tau) &= \arg \max_{\zeta} \prod_{\tau \in \mathcal{D}} \frac{\exp(\frac{1}{\beta} r(\tau)) \zeta(\tau)}{Z(\zeta)} \\
        \text{where }& Z(\zeta) = \int_{\tau} \exp(\frac{1}{\beta} r(\tau)) \zeta(\tau) \, d\tau.
    \end{aligned}
    \label{eq3:MLE_problem}
\end{equation}

$\zeta(\tau) = \prod_t \prod_i \zeta_i(s_t, a_t)$  can be decomposed into a product of feasibility factors $\zeta_i(s_t, a_t) : \mathcal{S} \times \mathcal{A} \rightarrow (0,1)$ for each state-action pair. The feasibility that each state-action pair for all $K$ constraints is denoted as $\zeta(s_t, a_t) = \prod_i \zeta_i(s_t, a_t)$. Note that $\zeta(s, a) \notin \{0, 1\}$ due to the properties of the sigmoid output. From here, we set $c(s, a)$ with $\bar{\zeta}(s, a) = 1 - \zeta(s, a): \mathcal{S} \times \mathcal{A} \rightarrow (0,1)$ and interpret $\zeta(\tau)$ as an estimate of the probability that $\tau$ will be feasible, similar to the case of soft constraints. In this context, the $\epsilon$ value is usually set just above zero, allowing it to operate like a hard constraint. When calculating the gradient of the constraint model to optimize the log-likelihood in Eq. \ref{eq3:MLE_problem}, the reward term can be ignored. The partition function $\log Z(\zeta)$ can also be estimated through a sample-based approximation using nominal policy $\pi$ learned in the forward step. Thus, the update of the constraint function is derived by matching the expected gradient of $\log \zeta$ for each expert and nominal trajectories as follows \cite{malik2021inverse}:

\begin{equation}
\mathbb{E}_{\tau_{E} \in \mathcal{D}_{E}} [\nabla_\zeta \log \zeta(\tau_{E})] - \mathbb{E}_{\tau \sim \pi} [\omega(\tau) \nabla_\zeta \log \zeta(\tau)],
\label{eq4:constraint_update}
\end{equation}

where $\omega(\tau)=\frac{\zeta}{\zeta'}$ represents importance sampling weights, defined as the ratio relative to $\zeta'$ computed in earlier iterations. Additionally, the subscript E indicates elements associated with the expert. Assuming that the expert policy $\pi_{E}$ is known, we have access to expert trajectories $\tau_{E}$ sampled from $\mathcal{D}_{E}=\{\tau_{E}\}_{j=1}^{N_{E}}$. The standard ICRL method uses an iterative approach that alternates between updating the policy and the constraint function based on Eq. \ref{eq1:CRL_problem} and Eq. \ref{eq4:constraint_update}, as illustrated on the right side of Fig. \ref{fig1a:stadard_icrl_architecture}. In practice, the constraint function is approximated using the complement of $\zeta$ obtained from the inverse step, treating it as $c_{i} \approx \bar{\zeta}_{i} = 1 - \zeta_{i}$ to establish the constraints. The nominal policy is then updated using the PPO Lagrange algorithm \cite{ray2019benchmarking}.

\section{Distribution-Informed Adaptive Learning}

\begin{figure}
    \centering
    \includegraphics[width=1.0\linewidth]{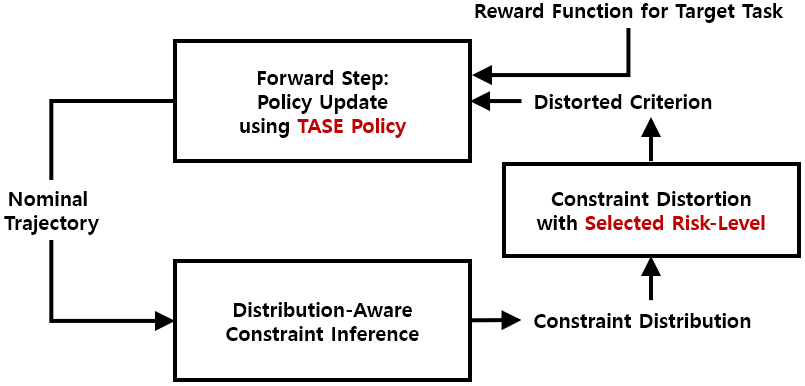}
    \caption{Architecture of safe TL stage in DIAL. The constraint model is used solely for inference at this stage, while only the policy is updated for TL. We highlight the advantages of the proposed method, which leverages the constraint distribution learned in the previous stage, safe IL, and the TASE policy, through the text marked in red.}
    \label{fig2:DIAL_SafeTL_architecture}
\end{figure}

In this paper, we aim to enable agents to safely adapt to diverse environments in navigation and control domains, particularly in safety-critical systems. To achieve this objective, we propose the DIAL method, a two-stage approach designed for environments represented by a CMDP. This method first performs safe IL and then shifts its focus to safe TL. The core idea of DIAL is to simultaneously learn a constraint function that is aware of the risk distribution across various tasks and a policy encouraging safe exploration of new tasks. These components are then used to facilitate safe adaptation. In the first stage, safe IL, we use multi-task demonstrations, as illustrated on the right side of Fig. \ref{fig1b:proposed_icrl_problem}, to learn both the constraint distribution and a TASE policy. Since the true constraints of the environment are unknown in this stage, we rely on expert demonstrations that accomplish multiple tasks. We assume that while these demonstrations meet all safety requirements in the original environment, they may be sub-optimal in adapted environments due to inherent biases in the expert’s risk preference. 
In the second stage, safe TL, as shown in Fig. \ref{fig2:DIAL_SafeTL_architecture}, we utilize the distorted criterion and TASE policy. Both are flexibly adjusted based on the risk level to enable safe and efficient adaptation to meta-tasks. To illustrate the benefits of DIAL in this stage, we configure a changed environment with the same safety requirements but a new target task. The underlying hypothesis of DIAL is that the TASE strategy with awareness of the constraint distribution is crucial for managing potential risks arising from limited data while facilitating adaptation to new tasks. Notably, DIAL can also be effectively coupled with a stable linear model-based controller \cite{leurent2019approximate} for autonomous driving, enabling structured exploration while maintaining lane alignment. The remainder of this section provides a detailed explanation of the design of DIAL. We begin by introducing methods to infer the constraint distribution from multi-task demonstrations that allow for flexible adjustment of the risk level. We also explain how to derive a policy that supports TASE. We then describe the process of carefully adjusting the learned constraint distribution by selecting an appropriate risk level to ensure safety in the changed environment and utilizing the TASE policy to facilitate meta-task resolution.

\subsection{Constraint Inference with Risk-Sensitive Criterion}

When designing the distribution-aware constraint in Fig. \ref{fig1b:proposed_icrl_architecture}, we aim to capture the diverse risk preferences in expert demonstrations that meet safety requirements, even for randomly assigned tasks. To achieve this, we employ a learning approach that infers distorted constraint distributions with properly selected risk levels. Instead of maximum likelihood estimation for $\zeta(\tau)$, Bayesian methods \cite{papadimitrioubayesian, gaurav2023learning, xu2023uncertainty} can estimate a posterior distribution $p(\zeta(\tau) \vert \mathcal{D})$ by combining a prior $p(\zeta(\tau))$ with the trajectory likelihood $p(\mathcal{D} \vert \zeta(\tau)) \approx \pi(\tau)$, as specified in Eq. \ref{eq2:MaxEnt_model}. Previous methods calculate the constraint $\mathbb{E}_{\tau \sim \pi(\cdot),\zeta \sim p(\cdot \vert \mathcal{D})} [\bar{\zeta}(\tau)] \leq \epsilon $ to ensure the mean of the estimated distribution remains below the threshold. However, this approach is vulnerable to long or heavy-tailed distributions, rendering it insufficiently stringent for handling rare but extreme situations. The mean constraint ultimately lacks the flexibility to adjust risk levels for changes in the unsafe set, shown as the green polygon in Fig. \ref{fig1b:proposed_icrl_problem}. This limitation becomes even more pronounced in multi-task settings, where the attraction region, marked by the blue boundary, expands to cover more tasks. Although a previous study \cite{xu2023uncertainty} addresses this issue by employing a risk-sensitive criterion like CVaR, it focuses solely on optimizing the policy for a single task during the forward step. In contrast, we emphasize using CVaR in the inverse step to learn constraints capable of handling scalable novel tasks.

\begin{figure}
    \centering
    \includegraphics[width=1.0\linewidth]{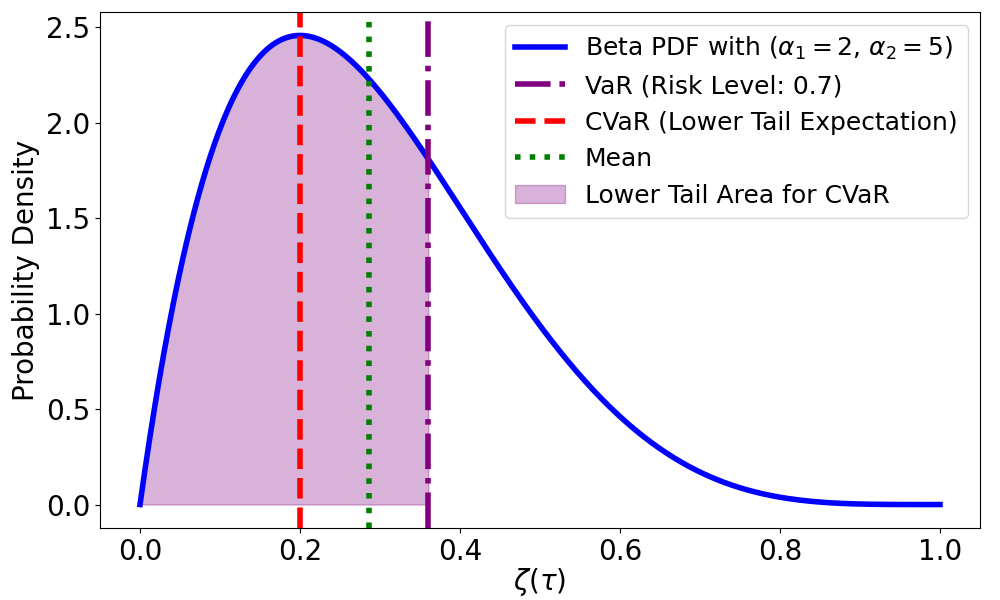}
    \caption{Difference between CVaR and mean in Beta distribution. We set $\alpha$ and the risk level $\lambda$ at fixed values for illustration purposes.}
    \label{fig3:risk-sensitive_criterion}
\end{figure}

When estimating statistical metrics like the mean or CVaR of a posterior distribution, it is often challenging because arbitrary distributions lack a precise closed form. Therefore, it is necessary to approximate them with more manageable distributions. In our case, we would like to model the probability that a trajectory is safe as a random variable that falls within a finite interval between 0 and 1. Consequently, we choose an ideal Beta distribution as the posterior distribution and set $ p(\zeta(\tau) \vert \mathcal{D}) \approx q(\zeta(\tau) \vert \alpha)$. The Beta distribution is particularly useful as it can represent asymmetric or heavy-tailed distributions. This distribution reflects the ratio of binary outcomes, such as successes and failures, of safe paths and is characterized by two parameters, $\alpha = [\alpha_{1}, \alpha_{2}]$. For simplicity in notation, we omit $i$ and assume a single constraint case. To handle multiple constraints, we can factorize $q(\zeta(\tau) \vert \alpha) = \prod_i q(\zeta_{i}(\tau) \vert \alpha^{i})$, expressing it as a product of independent components.

We use CVaR to measure distorted risk, taking into account the probability of a path being safe at a specified risk level within the Beta distribution. Fig. \ref{fig3:risk-sensitive_criterion} illustrates how CVaR is calculated in contrast to the mean. First, we determine the value at risk (VaR), which represents the threshold probability of safety at a given risk level $\lambda$. This value is the $\lambda$-percentile $F_{\zeta(\tau)}^{-1}(\lambda;\alpha)$, with $\text{VaR}_{\lambda}$ defined as the lower $\lambda$-quantile of the distribution:

\begin{equation}
\text{VaR}_{\lambda} = \inf \{ x \in (0, 1) : F_{\zeta(\tau)}(x;\alpha) \geq \lambda \},
\end{equation}

where $F_{\zeta(\tau)}(x;\alpha)$ represents the cumulative distribution function (CDF) of the $q(\zeta(\tau) \vert \alpha)$ distribution. Next, CVaR is obtained by calculating the expected value over the region below $\text{VaR}_{\lambda}$ as follows:

\begin{equation}
\text{CVaR}_{\lambda} = \mathbb{E}_{\zeta(\tau) \sim q(\cdot \vert \alpha)}[\zeta(\tau) | \zeta(\tau) \leq \text{VaR}_{\lambda}].
\end{equation}

This represents the average of extreme values at a given risk level. Note that if $\lambda = 1$, CVaR is equal to the mean. We employ a neural network $f_{\phi}(\alpha \vert \tau)$, which takes a trajectory $\tau$ as input and outputs the parameter $\alpha$ of a Beta distribution. This approach facilitates constraint learning by capturing the risk distribution and incorporating CVaR. Since the two parameters of the Beta distribution are positive, the final layer is implemented with a softplus activation. We interpret the network's output as variables sampled from two Gamma distributions, treating the approximated network as a prior model for the Beta distribution. This approach allows the shape of the Beta distribution to be determined by random variables generated by the Gamma distributions, rather than fixed parameters, thereby capturing uncertainty in the data. This flexibility makes probabilistic modeling more adaptable and relevant to the context. Therefore, we update the constraint model by applying the risk-sensitive criterion $\Gamma_{\phi}^{\lambda}(\tau) \doteq  \mathbb{E}_{\alpha \sim f_{\phi}(\cdot \vert \tau)} [\text{CVaR}_{\lambda}]$ in place of $\zeta(\tau)$ in Eq. \ref{eq4:constraint_update}. Consequently, the gradient of the loss $\nabla_\phi \mathcal{L}_{C}(\phi, \lambda)$ is formulated as:

\begin{equation}
\mathbb{E}_{\tau_{E} \in \mathcal{D}_{E}} [\nabla_\phi \log \Gamma_{\phi}^{\lambda}(\tau_{E})] - \mathbb{E}_{\tau \sim \pi} [\omega(\tau) \nabla_\phi \log \Gamma_{\phi}^{\lambda}(\tau)].
\label{eq7:proposed_constraint_loss}
\end{equation}

In addition, considering diverse risk levels is known to be beneficial for acquiring risk-sensitive knowledge \cite{majumdar2017risk}. For this reason, we update the network using risk level $\lambda$ sampled from the uniform distribution $\mathcal{U}(0,1)$ in the safe IL stage where multi-task learning is conducted. In the later safe TL stage, where achieving high performance on the target task is crucial, fine-tuning is performed using grid search to determine $\lambda$. Following this, we include a regularization term when optimizing the evidence lower bound (ELBO) for the approximate posterior as follows:

\begin{equation}    
    \mathcal{L}_{P}(\phi) = \mathbb{E}_{\tau \sim \{\mathcal{D}_{E}, \mathcal{D}\}}\big[\mathcal{D}_{KL}[q\big(\zeta \vert f_{\phi}(\alpha \vert \tau)\big) \parallel p(\zeta)]\big]    
    \label{eq8:constraint_regularizer}    
\end{equation}

Given that both distributions $q(\cdot)$ and $p(\cdot)$ are Beta distributions, the Kullback-Leibler (KL) divergence can be computed in closed form, as shown in \cite{liu2023benchmarking}.

\subsection{Policy Improvement with Task-Agnostic Safe Exploration}

We introduce a novel policy improvement method within the ICRL framework that ensures safety by allowing limited exploration at an acceptable risk level, enabling effective learning across diverse tasks. Through this proposed approach, we obtain a TASE policy that helps in learning shared knowledge across various tasks. This enables the recovery of a more flexible constraint function that supports robust inference by accounting for varying risk levels and task-specific environmental changes. Our approach builds on the observation that using data spanning a wider range of tasks improves the accuracy of generalizable constraint learning \cite{kim2024learning}. Unlike prior approaches that either set constraints tailored to specific tasks \cite{subramanian2024confidence} or apply the same constraints across all tasks \cite{lindner2024learning}, our method provides improved generalization and adaptability to new situations.

Learning a policy in IL involves aligning it to frequently visit the state-action spaces observed in the demonstrations. Effective training requires the nominal policy to aim for an even coverage of a wide range of states. This approach enables divergence in the probability density of state-action coverage from the expert policy, thereby facilitating learning. In practice, this approach involves increasing the entropy of the probability density over the state-action pairs that the agent visits. To achieve this, Eq. \ref{eq1:CRL_problem} includes an entropy regularizer term $\mathcal{H}(\pi)$ that directly measures the probability distribution $\pi(\tau)$ of the nominal policy, thereby enhancing exploration. However, this naive method has limitations as it focuses solely on the probability distribution of the policy without capturing the correlations between states. In addition, simply relying on random actions for exploration can be inefficient, particularly in high-dimensional spaces or complex tasks. When the policy repeatedly selects high-reward actions early in training, confidence in these actions increases, resulting in lower entropy and reduced exploration. This can eventually hinder learning or prevent the achievement of goals. 

To address these limitations, we promote structured exploration by incorporating correlations between states into our entropy estimation. This approach is implemented under constraints that ensure safety. By focusing on task-relevant states, we interpret the increase in entropy as an exploration bonus in the policy update process, effectively enhancing task entropy. To extend to complex domains, we can indirectly model the state density function $\rho : \mathcal{S} \rightarrow \mathbb{R}_{\geq0}$ using $M$ particle groups $\mathcal{S} = \{s_{i}\}_{i=1}^{M}$ \cite{singh2003nearest}. Using these particle groups, we represent the average state density visited by the policy $\pi$ as $\rho_{\pi}(s) = \frac{1}{T}\sum_{t=0}^{T}\gamma \rho(s_{t}=s \vert \pi)$ for time horizon $T$, where $\int_{\mathcal{S}} \rho_{\pi}(s) \, \text{d}s = 1$. The entropy of the state density $\rho_{\pi}$ is computed as $\mathcal{H}(\rho_{\pi}) = - \int_{s \in \mathcal{S}} \rho_{\pi}(s) \ln \rho_{\pi}(s) \, ds$. In practice, we approximate this entropy using the particle groups, yielding $-\sum_{i=1}^{M} \hat{\rho}_{\pi}(s_{i}) \ln \hat{\rho}_{\pi}(s_{i}) \Delta s_{i}$, where $\Delta s_{i}$ represents the interval width around each particle $s_{i}$. The approximated density $\hat{\rho}_{\pi} = \frac{k}{M \cdot V_i^k}$ for these particles is determined by estimating the volume of a hypersphere formed by the radius $R_{i} = \vert x_{i} - x_{i}^{k\text{-NN}}\vert$. This calculation uses the $k$-NN method for each particle, where $x_{i}^{k\text{-NN}}$ represents the position of the $k$-th nearest neighbor ($k$-NN) particle to $x_{i}$. This density gives the $k$-NN entropy estimator \cite{ajgl2011particle} as follows:

\begin{equation}
    \begin{aligned}
    \hat{\mathcal{H}}_{k}(\rho_{\pi}) &= - \frac{1}{M}\sum_{i=1}^{M} \ln \frac{k}{M V_{i}^{k}} + \ln k - \Psi(k) \\
    &\text{where } V_{i}^{k} = \frac{R_{i}^{|\mathcal{S}|} \pi^{|\mathcal{S}| / 2}}{\Gamma \left(\frac{|\mathcal{S}|}{2} + 1\right)}.
    \end{aligned}
    \label{eq9:kNN_estimator}
\end{equation}

Here, $\Gamma(\cdot)$ represents the gamma function, and $\Psi(\cdot)$ is the digamma function, which is the logarithmic derivative of the gamma function. The last two terms correct the average bias introduced to address entropy underestimation for small $k$ values. To integrate this entropy estimator into RL, we use samples from the old policy $\pi_{\bar{\theta}}$ to approximate the state entropy of the current policy $\pi_{\theta}$. The policy $\pi$ is parameterized by a neural network with parameters $\theta$. For simplicity, we denote $\rho_{\pi_{\theta}}$ as $\rho_{\theta}$, omitting $\pi$ in the notation. Subsequently, to handle the discrepancy between the sampling policy $\pi_{\bar{\theta}}$ and the target policy $\pi_{\theta}$, we apply an importance-weighted (IW) $k$-NN estimator \cite{mutti2021task} as follows:

\begin{equation}
    \begin{aligned}
    \hat{\mathcal{H}}_{k}(\rho_{\theta} \vert \rho_{\bar{\theta}}) &= - \sum_{i=1}^{M} \frac{W_{i}}{k} \ln \frac{W_{i}}{V_{i}^{k}} + \ln k - \Psi(k) \\
    \text{where } W_{i} &= \sum_{j \in \mathcal{N}_{i}^{k}} w_{j}, \\
    \text{and } w_{j} &= \frac{\rho_{\theta}(x_{j}) / \rho_{\bar{\theta}}(x_{j})}{\sum_{n=1}^{M} \rho_{\theta}(x_{n}) / \rho_{\bar{\theta}}(x_{n})}.
    \end{aligned}
    \label{eq10:IW_kNN_estimator}
\end{equation}

Here, $\mathcal{N}_{i}^{k}$ is the set of indices of $k$-NN of $x_{i}$. In this context, when $\theta = \bar{\theta}$ and a uniform weight $w_{j} = \frac{1}{N}$is applied, we obtain $\hat{\mathcal{H}}_{k}(\rho_{\theta} \vert \rho_{\bar{\theta}})  = \hat{\mathcal{H}}_{k}(\rho_{\theta})$. Notably, in the IW $k$-NN estimation approach, trajectories are sampled independently, while the states within each trajectory account for the correlations among neighboring particles. Furthermore, to stabilize the convergence, we utilize a KL estimator $\hat{\mathcal{D}}_{KL}$ \cite{yang2023cem}. This is computed as the difference between the IW $k$-NN estimator in Eq. \ref{eq10:IW_kNN_estimator} and the estimator in Eq. \ref{eq9:kNN_estimator}, with the bias correction term canceling out. As long as the updated policy satisfies $\hat{\mathcal{D}}_{KL}[\rho_{\theta} \parallel \rho_{\bar{\theta}}] \leq \delta$, we can optimize the policy multiple times, serving as a trust-region constraint. 

Eventually, we can replace the constraint with the risk-sensitive criterion and the entropy regularizer term with the IW k-NN estimator in Eq. \ref{eq1:CRL_problem}. We then solve an unconstrained min-max optimization problem by applying the Lagrangian method to the objective function \cite{tessler2018reward}. This approach allows us to handle the original constrained problem as an equivalent unconstrained problem, which we define as follows:

\begin{equation}
    \min_{\kappa \geq 0} \max_{\theta} \mathcal{J}_{R}(\theta) + \beta \mathcal{J}_{H}(\theta) - \kappa \mathcal{J}_{\kappa}(\phi, \lambda),
    \label{eq11:policy_objective}
\end{equation}

where $\mathcal{J}_{R}(\theta) = \mathbb{E}_{\tau \sim \pi_{\theta}(\cdot)} \left[r(\tau) \right]$ encourages the maximization of expected reward-return. Moreover, $\mathcal{J}_{H}(\theta) = \hat{\mathcal{H}}_{k}(\rho_{\theta} \vert \rho_{\bar{\theta}})$ is an entropy-based term, scaled by $\beta$ to control the degree of entropy regularization, thereby encouraging exploration in the policy. Lastly, $\mathcal{J}_{\kappa}(\phi, \lambda) =  \mathbb{E}_{\tau \sim \pi_{\theta}(\cdot)}[\bar{\Gamma}_{\phi}^{\lambda}(\tau)] - \epsilon$ is a constraint term that ensures the policy remains within the constraint threshold with an expected risk $\bar{\Gamma}_{\phi}^{\lambda} = 1 - \Gamma_{\phi}^{\lambda}$. In this setup, $\kappa$, also referred to as the safety weight, is a Lagrange multiplier associated with the constraint term $\mathcal{J}_{\kappa}$.
By optimizing this Lagrangian formulation, we effectively balance maximizing rewards and entropy while minimizing the constraint violation. Algo. \ref{algo1:safeIL} and \ref{algo2:safeTL} summarize our training procedure for the safe IL and safe TL stages in DIAL, respectively.

\begin{figure}[!th]
    \centering
        \begin{algorithm}[H]  
        \caption{Safe IL in DIAL}
            \label{algo1:safeIL}
            \begin{algorithmic}[1]       

                \State \textbf{Given}: Entropy coefficient $\beta$, budget $\epsilon$, learning rates for $\eta_{C}$, $\eta_{P}$, and $\eta_{\kappa}$,  expert trajectories $\mathcal{D}_{E}$, number of neighbors $k$, and trust-region threshold $\delta$ 
                \State \textbf{Initialize}: Network parameters $\theta$, $\phi$, and safety weight $\kappa$
                
                \For{each epoch \do}      
                
                    \State Rollout buffer $\mathcal{D} \gets \emptyset$ 
    
                    \For{each environmental step \do}
                        \State Execute action $a \sim \pi_{\theta}(a \vert s)$
                        
                        \State Observe next state $s' \sim \mathcal{P}(s' \vert s, a)$
    
                        \State Add transition to buffer $\mathcal{D} \gets \mathcal{D} \cup \{(s, a, s')\}$
                        
                        
    
                    \EndFor
                    
                    \For{each gradient step \do}
    
                        \State Sample $\tau_{E} \sim \mathcal{D}_{E}$ and $\tau \sim \mathcal{D}$, respectively      
    
                        \State Sample risk level $\lambda \sim \mathcal{U}(0,1)$                                              

                        \State Update constraint $f_{\phi}(\alpha \vert \tau)$ with Eq. \ref{eq7:proposed_constraint_loss} and \ref{eq8:constraint_regularizer}:
    
                        \State \hspace{1.0em} $\phi \gets \phi + \eta_{C} \nabla_\phi \mathcal{L}_{C}(\phi, \lambda) - \eta_{P} \nabla_\phi \mathcal{L}_{P}(\phi)$

                        
                        \State Update policy $\pi_{\theta}(a \vert s)$ with Eq. \ref{eq11:policy_objective}:       
    
                        \State \hspace{1.0em} $\kappa \gets \max(0, \kappa + \eta_{\kappa} \mathcal{J}_{\kappa}(\phi, \lambda))$
    
                        \While{Trust-region estimator $\hat{\mathcal{D}}_{KL} \leq \delta$}
                            \State $\theta \gets \theta - \beta \nabla_{\theta} \mathcal{J}_{H}(\theta)$
                        \EndWhile                        
                    \EndFor
                \EndFor
            \end{algorithmic}
        \end{algorithm}
\end{figure}

\begin{figure}[!th]
    \centering
        \begin{algorithm}[H]  
        \caption{Safe TL in DIAL}
            \label{algo2:safeTL}
            \begin{algorithmic}[1]       
                \State \textbf{Given}: Entropy coefficient $\beta$, budget $\epsilon$, learning rates for $\eta_{R}$ and $\eta_{\kappa}$, target reward function $r$, and risk level $\lambda$ 
                \State \textbf{Initialize}: Parameters of networks $\theta$ and $\phi$ from Safe IL, and safety weight $\kappa$
                
                \For{each epoch \do}      
                
                    \State Rollout buffer $\mathcal{D} \gets \emptyset$ 
    
                    \For{each environmental step \do}
                        \State Execute action $a \sim \pi_{\theta}(a \vert s)$
                        
                        \State Observe next state $s' \sim \mathcal{P}(s' \vert s, a)$

                        \State Add transition to buffer $\mathcal{D} \gets \mathcal{D} \cup \{(s, a, r, s')\}$
                                                
                    \EndFor
                    
                    \For{each gradient step \do}
    
                        \State Sample $\tau \sim \mathcal{D}$       
                        \State Recover constraint $\hat{c}(\tau) \gets \bar{\Gamma}_{\phi}^{\lambda}(\tau)$
                        
                        \State Update policy $\pi_{\theta}(a \vert s)$ with Eq. \ref{eq1:CRL_problem}:       
    
                        \State \hspace{1.0em} $\kappa \gets \max\big(0, \kappa +  \eta_{\kappa} (\mathbb{E}_{\pi_{\theta}(\tau)} \left[ \hat{c}(\tau) \right] - \epsilon)\big)$

                        \State \hspace{1.0em} $\theta \gets \theta - \eta_{R}\nabla_{\theta} \mathbb{E}_{\pi_{\theta}(\tau)} \left[r(\tau) \right] - \beta\nabla_{\theta} \mathcal{H}(\pi_{\theta}(\tau))$

                    \EndFor
                \EndFor
            \end{algorithmic}
        \end{algorithm}
\end{figure}

\section{Experiments}

We organize the experimental analysis using the following two aspects to evaluate our proposed method. First, we demonstrate our method that encourages safe exploration at varying risk levels by using multi-task demonstrations instead of explicit constraints in safe IL. Second, we reveal that leveraging our recovered risk-sensitive constraints and safe exploration policy can accelerate learning of the target task and show benefit safety assurance in safe TL.

\subsection{Benchmarks} 

Safe IL and TL are evaluated within the navigation and control domain, encompassing scenarios with risky situations involving static or dynamic obstacles across state spaces ranging from low to high dimensions. In safe IL, the aim is to perform exploration that navigates through as many safe states as feasible for arbitrarily given tasks. In safe TL, the objective is to conduct safe exploitation that effectively tackles a given target task while ensuring it remains within the constraints. To demonstrate the advantages of these two components, we have composed urban driving tasks that address real-world navigation and robot control tasks to validate performance across diverse environments, further details are described in the following paragraphs.

\begin{figure*}[!ht]
    \centering        
    \subfloat[Multi-Task Scenario]{        
        \centering
        \begin{minipage}{0.3\textwidth}             
            \includegraphics[width=1.0\textwidth]{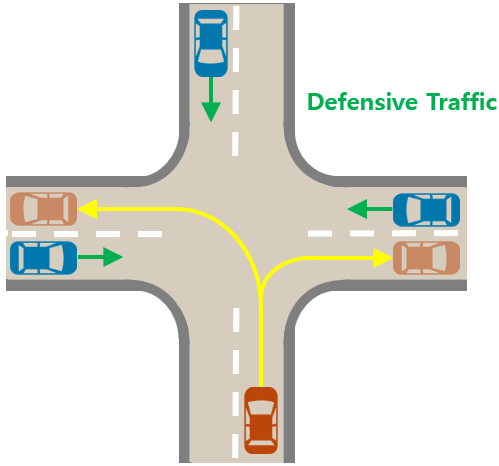} 
            \label{fig4a:multi-task_driving}
        \end{minipage}        
    }\hfill
    \subfloat[Shared Constraints]{        
        \centering
        \begin{minipage}{0.3\textwidth}
            \includegraphics[width=1.0\textwidth]{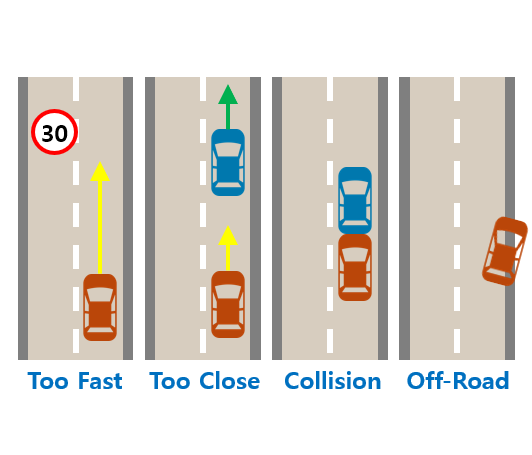} 
            \label{fig4b:driving_constraints}
        \end{minipage}
    } \hfill
    \subfloat[Meta-Task Scenario]{        
        \centering
        \begin{minipage}{0.3\textwidth}
            \includegraphics[width=1.0\textwidth]{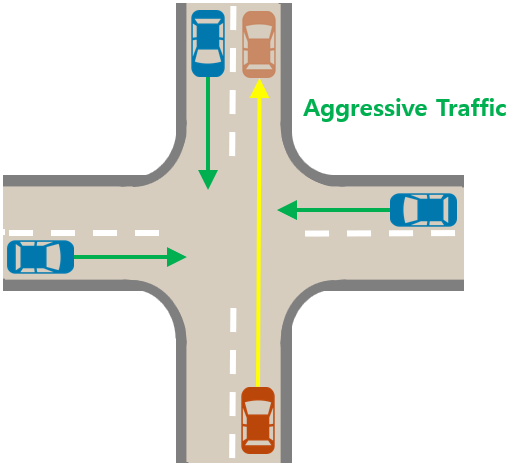} 
            \label{fig4c:meta-task_driving}
        \end{minipage}
    }
    \caption{Unsigned intersection environments in urban driving. The agent controls a red car, guiding it toward its destination by following a yellow arrow, while the surrounding blue cars are set to follow arbitrary paths at a fixed speed, indicated by green arrows. The goal of this environment is for the agent to learn the shared constraints (b) across left and right turns from the provided data in scenario (a). Then, using the learned constraints without additional data, the agent aims to safely reach a new destination in changed scenario (c), even with aggressive traffic flows.}
    \label{fig4:urban_driving}
\end{figure*}

\textbf{Urban Driving:} To compare environments characterized by dynamic risks in high-dimensional states, we use a modified version of the intersection environment provided by HighwayEnv \cite{leurent2018environment}, referring \cite{lindner2024learning}. Fig. \ref{fig4:urban_driving} illustrates safe IL, where the agent learns shared constraints (\ref{fig4b:driving_constraints}) from demonstrations in the multi-task scenario (\ref{fig4a:multi-task_driving}), and safe TL, where the agent applies this learned knowledge to adapt in the meta-task scenario (\ref{fig4c:meta-task_driving}). The observation space includes 7 types of kinematic information, such as position and speed, for the agent and 15 surrounding vehicles. To achieve a permutation-invariant representation of the $\mathbb{R}^{15 \times 7}$ input independently of the order of surrounding vehicles, we use an attention-based encoder as the backbone network for the constraint model $f_{\phi}(\alpha \vert \tau)$. This encoder captures the relative importance of each surrounding vehicle and calculates a weighted sum to represent the input as a latent variable in $\mathbb{R}^{32}$. We adopt the same controllers, reward function, and constraints as specified in \cite{lindner2024learning} to ensure fair comparisons with the baselines. Each vehicle follows a fixed path by adjusting acceleration and steering angle through a linearized controller parameterized in $\mathbb{R}^{5}$ \cite{leurent2019approximate}. Surrounding vehicles maintain fixed parameters, while the agent’s parameters are optimized using a constrained cross-entropy method (CEM) \cite{wen2018constrained} rather than the PPO Lagrangian in Eq.\ref{eq11:policy_objective}. This approach addresses multiple constraints by first ranking parameters according to the number of constraint violations, then assessing them based on violation magnitude and reward. 
The reward function is designed as a linear combination of random variables and features to reflect various driving preferences, including reaching the target, driving speed, and lane changes:

\begin{equation}
10 \times \mathbb{1}_{\text{Goal}} + \xi_{v} v_{t} + \xi_{\angle}\vert \angle_{t}\vert,
\end{equation}

where $v_{t}$ represents the vehicle's speed, $\angle_{t}$ indicates the difference angle between the heading of the vehicle and the target lane, and $\xi_{v} \sim \mathcal{N}(0.1, 0.1)$ and $\xi_{\angle} \sim \mathcal{N}(-0.2, 0.1)$ are random variables. The target lane is the path extending to the given target location along the road network provided by the environment. The constraint function limits dangerous events by defining features composed of four metrics, $\varphi(s_{t},a_{t}):\mathcal{S}\times \mathcal{A} \rightarrow \{0,1\}^{4}$. These features include cases where the speed exceeds 15, the distance to the vehicle ahead is less than 10, a collision occurs, and the vehicle departs from the road. A safe situation is defined as one where the probability of each event occurring in an episode remains below the ground truth constraint thresholds $\epsilon = [0.2, 0.2, 0.05, 0.1]$. In this problem, we consider $1 - \max(0, \frac{1}{T}\sum_{t}^{T}\varphi(s_{t},a_{t}) - \hat{\epsilon})$ as the feasibility function and learn to infer thresholds $\hat{\epsilon}$ for the four defined features.

\begin{figure*}[!ht]
    \centering        
    \subfloat[MountainCar]{        
        \centering        
        \includegraphics[width=0.24\textwidth]{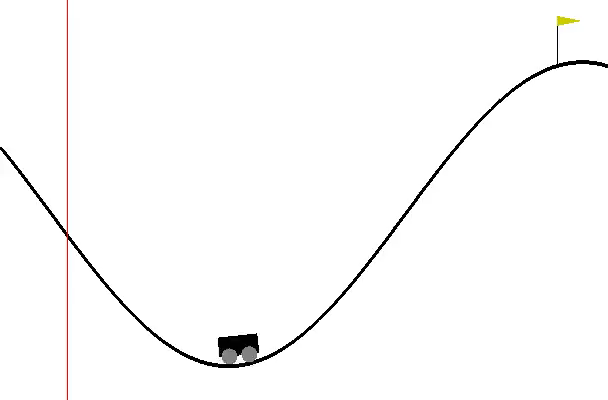} 
        \label{fig5a:mountaincar}        
    }
    \subfloat[CartPole]{        
        \centering        
        \includegraphics[width=0.24\textwidth]{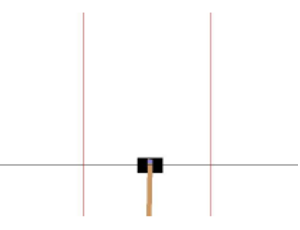} 
        \label{fig5b:cartpole}        
    }
    \subfloat[BasicNav]{        
        \centering        
        \includegraphics[width=0.24\textwidth]{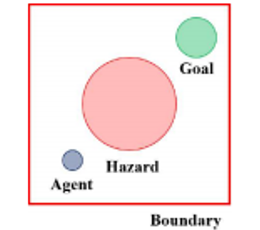} 
        \label{fig5c:basicnav}        
    }
    \subfloat[PointGoal]{        
        \centering        
        \includegraphics[width=0.24\textwidth]{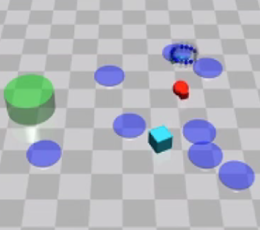} 
        \label{fig5d:pointgoal}        
    }
    \caption{Robot control environments that aim to perform target tasks while ensuring safety.}
    \label{fig5:robot_control}
\end{figure*}

\textbf{Robot Control:} To compare environments with static risks in low-dimensional states, we use a modified version of OpenAI Gym \cite{brockman2016openai}, shown in Fig. \ref{fig5:robot_control}a-c. Each environment has a safety area marked with a red line. For MountainCar, the goal is to reach the point where the right flag is located without the car going to the left of the red line. However, the agent is penalized so that a large action $a_{t}$ is not performed. For CartPole, the goal is to raise the pole angle $\theta_{t}$ vertically while keeping the cart inside both red lines. For BasicNav, the goal is to reduce the goal distance $d_{t} = s_{goal} - s_{t}$ while avoiding the circular hazard region in the center. For evaluating responses to risks involving randomness in high-dimensional states, we use Safety Gym \cite{ray2019benchmarking}, illustrated in Fig. \ref{fig5d:pointgoal}. For PointGoal, we aim to control the point robot to reach a random goal described as a green cylinder. In this environment, the agent is restricted from entering the blue circles depicted on the ground or pushing the vase marked with a cyan block. A vase moving upon contact and stationary blue circles are randomly generated within a certain range around the target point. 

The rewards and costs used to train the RL agent are given the prefix "extrinsic" for those provided by the environment and "intrinsic" for those recovered through IL. Each environment provides an extrinsic cost whenever unsafe interactions occur, but this is used only for performance evaluation and not for training. However, the extrinsic rewards from the environment and the recovered intrinsic costs are both used for training. Further details on each environment are provided in the Appendix. \ref{appendixA:environmental_settings}.

\subsection{Baselines}

Our approach, DIAL, is an algorithm designed to learn constraints that ensure safety when addressing new meta-tasks. These tasks are set in similar but slightly different environments from those demonstrated in multi-task settings. Several approaches based on IRL or ICL serve as natural starting points to address our constraint learning problem. This makes them suitable for comparison with our proposed method. We adopt the following three methods as baselines: MERL \cite{ziebart2008maximum}, MECL \cite{malik2021inverse}, and COCL \cite{lindner2024learning}. MERL and MECL both belong to the MaxEnt-based IL family. They can derive the optimal policy that imitates the expert from single-task demonstrations. The key difference between these methods lies that MERL recovers rewards, while MECL focuses on recovering constraints under the assumption that rewards are known. In MERL, the average of the inferred individual rewards for each task demonstration, $\sum_{i}^{N} \hat{r}_i$, can be added as a penalty term to the reward $r_{\text{eval}}$ for a new meta-task. In MECL, the average of the inferred individual constraints across multi-task demonstrations, $\sum_{i}^{N} \hat{c}_i$, is incorporated as a constraint term. This term is considered separately from the reward $r_{\text{eval}}$, under the assumption that the rewards for multi-task demonstrations are already known. COCL does not belong to the IL family that cannot obtain the policy. However, it can recover a shared constraint $\hat{c}$ even without knowing the individual rewards $r_i$ for each task. This is achieved by constructing the convex hull of the safety set based on feature expectations from the demonstration data. Subsequently, if \(r_{\text{eval}}\) and \(\hat{c}\) are known, a policy can be obtained by optimizing Eq. \ref{eq1:CRL_problem}. In the robot control environment illustrated in Fig. \ref{fig5:robot_control}, COCL is excluded as a baseline because it cannot be applied due to the lack of directly defined feature vectors for the state. We highlight that our approach, like COCL, learns constraints across multi-task demonstrations. However, DIAL can derive the TASE policy and considers the distribution of constraints. These aspects distinguish our method and provide notable advantages, particularly in helping to safely adapt to changed environments when addressing new tasks.

\begin{figure*}[!ht]
    \centering      
    \begin{minipage}{1.0\textwidth}               
        \centering
        \includegraphics[width=0.5\textwidth]{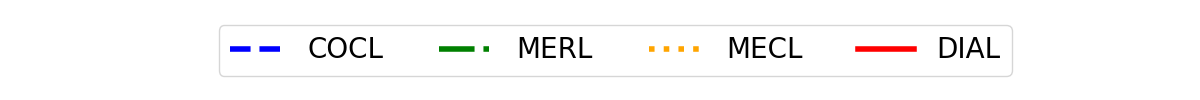}    
        \vspace{-0.5cm}
    \end{minipage}
    \subfloat[Multi-Task Learning with Safe IL]{   
        \begin{minipage}{0.48\textwidth}             
            \includegraphics[width=0.49\linewidth]{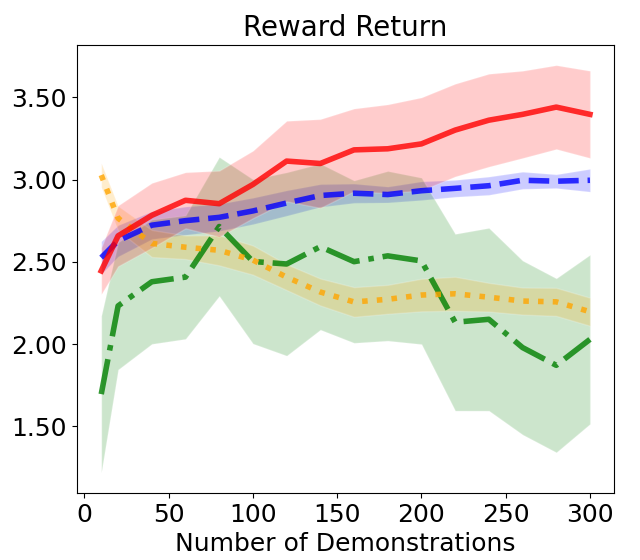}       
            \hfill
            \includegraphics[width=0.49\linewidth]{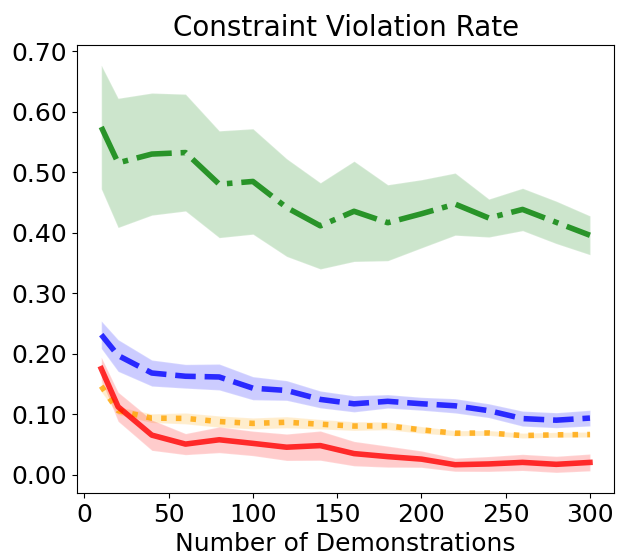}      
        \end{minipage} 
        \label{fig6a:urban_driving_results}
    }
    \hfill
    \subfloat[Meta-Task Learning with Safe TL]{
        \begin{minipage}{0.48\textwidth}             
            \includegraphics[width=0.49\linewidth]{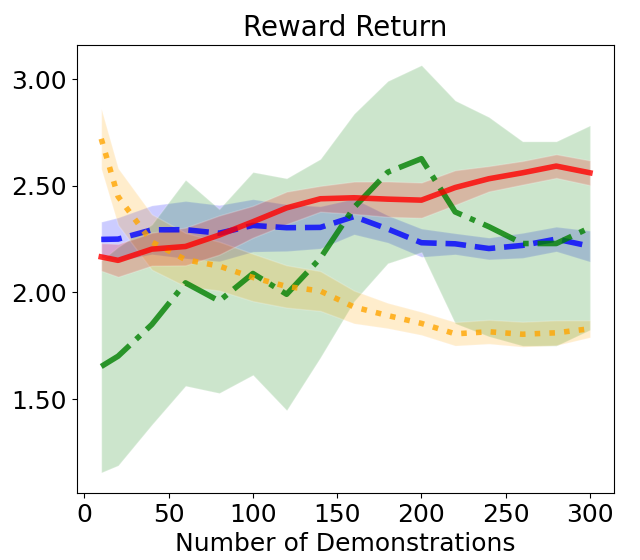}        
            \hfill
            \includegraphics[width=0.49\linewidth]{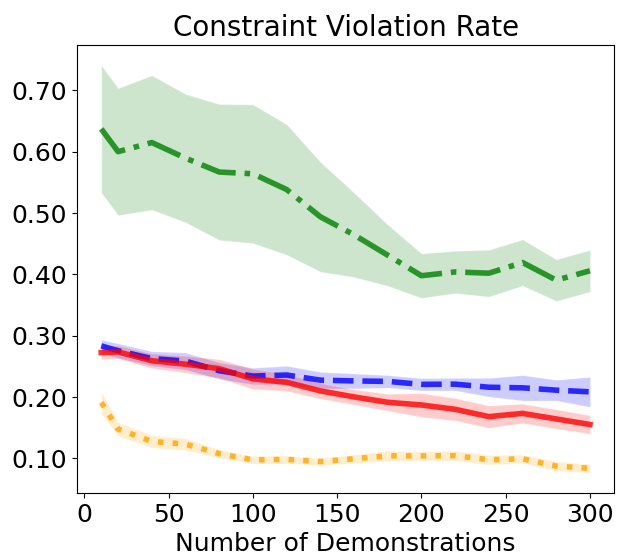} 
        \end{minipage} 
        \label{fig6b:urban_driving_results}
    }       
    \caption{Comparison of RR and CV for urban driving tasks based on the number of expert trajectories.}
    \label{fig6:urban_driving_results}
\end{figure*}

\begin{figure}[!ht]
    \centering
    \includegraphics[width=1.0\linewidth]{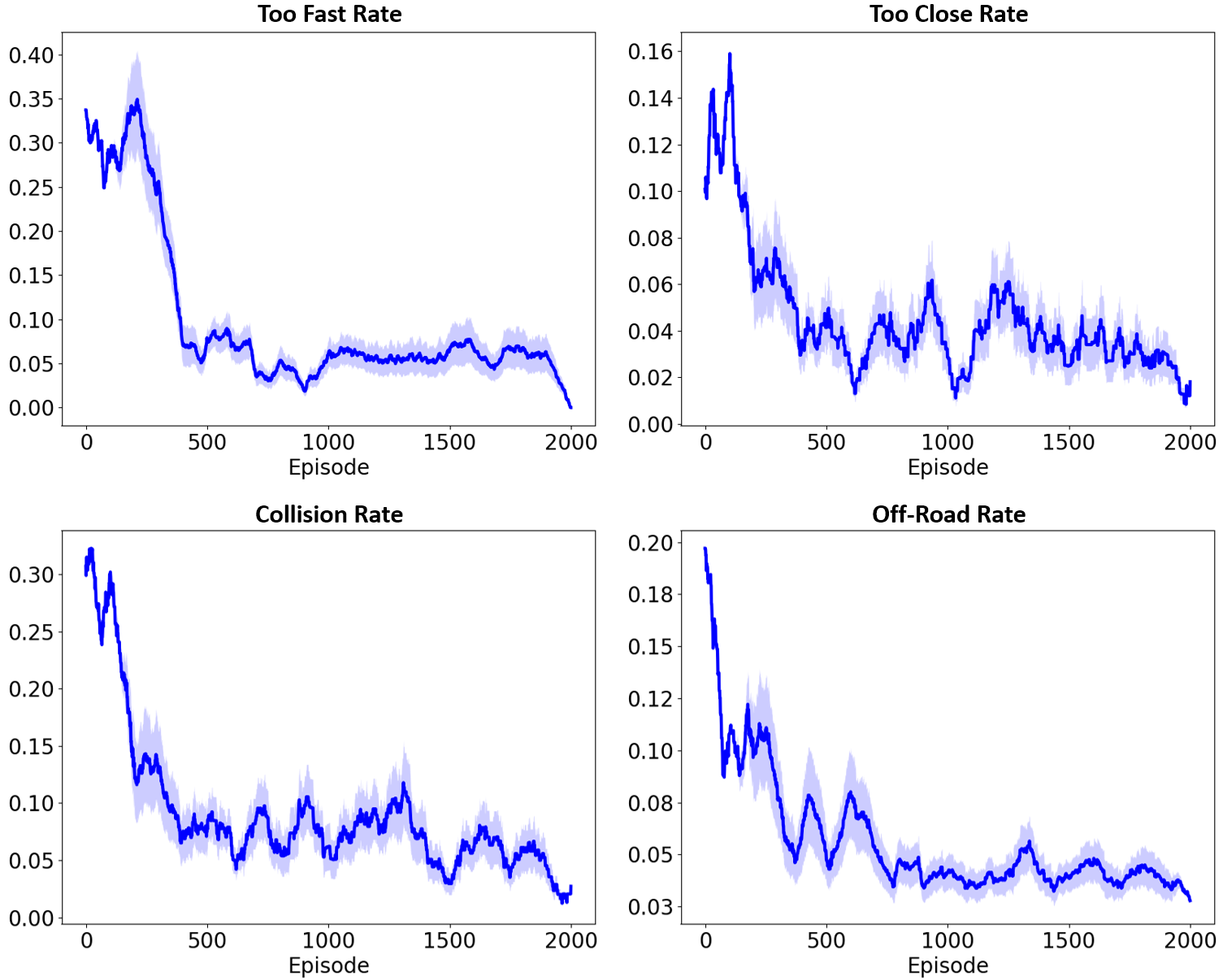}  
    \caption{Learning curves of DIAL for each constraint during safe IL.}
    \label{fig7:multi_constraints_learning_curves}
\end{figure}

\subsection{Implementation Details}

DIAL trains two learnable models: the constraint function and the policy. Both models have a two-layer neural network architecture with 256 hidden units and ReLU activation across all methods. The policy handles continuous action spaces by outputting the mean and variance of a Gaussian distribution. The mean is constrained to a specific range using a tanh output, while the variance ensures positive values using a softplus output. In our approach, the constraint function approximates the parameters of a Beta distribution, which is why we use a softplus output instead of softmax. All neural network parameters are updated using the Adam \cite{kingma2014adam} optimizer. As an exception, in the Urban driving environment, we employ an encoder that embeds inputs into permutation-invariant representations as the backbone model for the constraint function in all methods. Additionally, in this environment, the policy is represented by a linearized controller instead of neural network and is optimized using CEM \cite{wen2018constrained}. The hyperparameters used in the experiments were tuned through a coarse grid search. Appendix. \ref{appendixB:implementation_details} provides detailed descriptions of the expert demonstration collection method, hyperparameter selection, and stabilization techniques for training to facilitate experiment reproduction. We observed experimentally that DIAL achieves asymptotic performance within approximately 150K environmental steps in urban driving of Fig. \ref{fig4:urban_driving} and within 20K, 300K, 300K, and 1M environmental steps in robot control of Fig. \ref{fig5:robot_control}. We believe these results are due to the use of a distribution-aware constraint function, which allows for flexible adjustment of risk levels and cautious exploration of new tasks. All experiments were conducted on a PC with an Intel Xeon Gold 6248R CPU (3.00GHz, 48 cores), an NVIDIA GeForce RTX 3090 GPU, and 256GB of memory.

\begin{figure*}[!ht]
    \centering
    \includegraphics[width=1.0\linewidth]{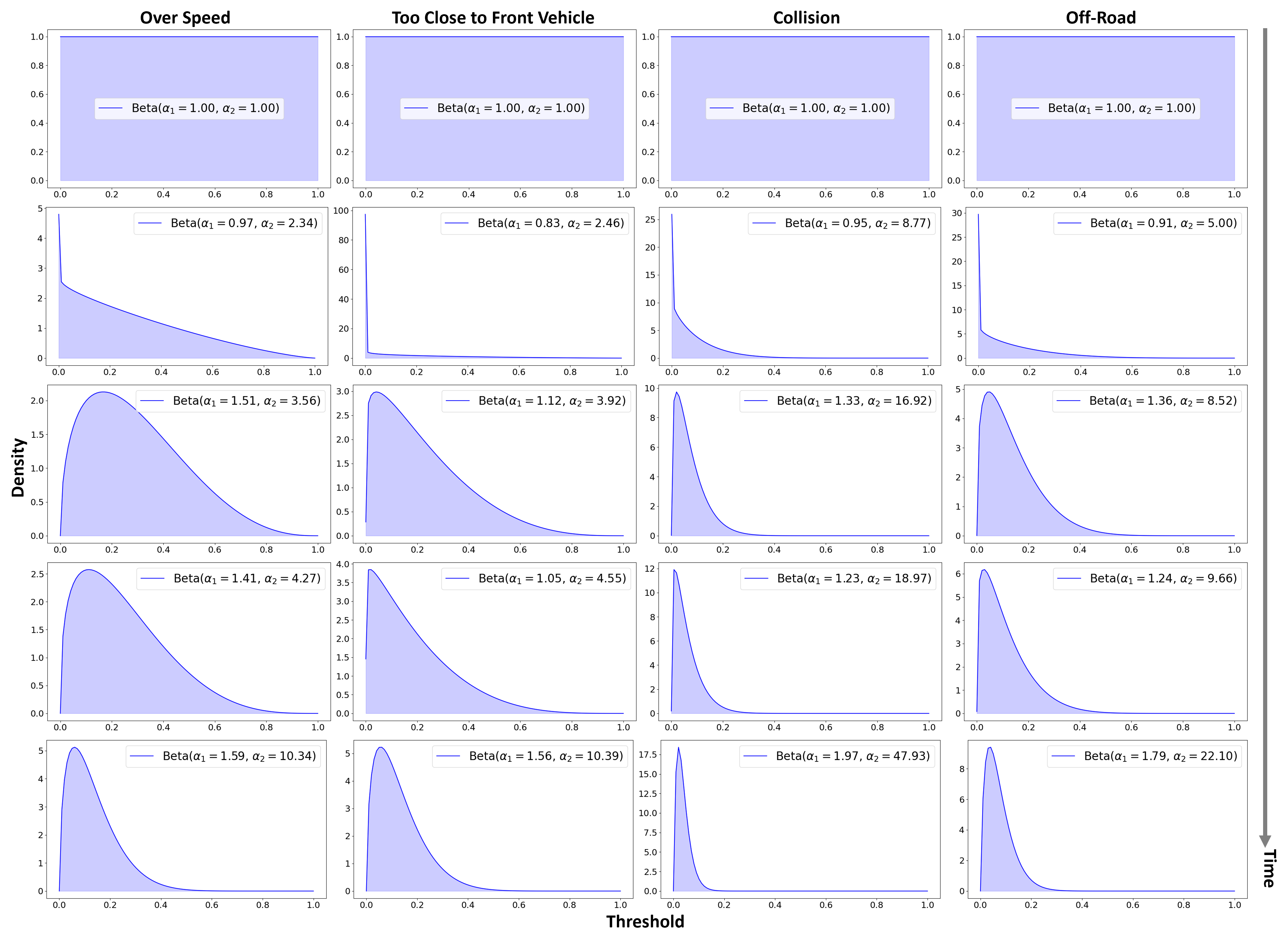}  
    \caption{Visualization of the changes in the constraint distribution for DIAL during safe IL.}
    \label{fig8:constraint_distribution}
\end{figure*}

\subsection{Metrics}
We use the following metrics to evaluate performance: reward-return (RR), cost-return (CR), constraint violation rate (CV), and state entropy (SE). RR and CR represent the average of the total explicit rewards and costs obtained by the agent during an episode, averaged across multiple episodes. RR reflects task performance, while CR evaluates safety performance. CV indicates the proportion of episodes where the value of CR divided by $T$ exceeds the specified constraint budget $\epsilon$, representing the likelihood of violating constraints in a single episode. When handling multiple constraints, the sum of all CVs is used. SE is calculated by discretizing two primary states in each environment and measuring the frequency with which the policy visits each state during all episodes. This metric is used to compare the exploration level of different policies. In particular, the map that assigns visit frequencies or inferred constraints to each discretized state is useful for qualitative comparisons. The two primary states are empirically selected for each environment to clearly highlight differences. For more details, please refer to Tab. \ref{table:environmental_settings} in Appendix. \ref{appendixA:environmental_settings}. All metrics are presented as the average values measured over 20 episodes for 5 seeds. The shaded areas in the plots represent the standard deviation.

\subsection{Results for Safe Imitation Learning}

This section evaluates the constraint function related to safety requirements and the policy associated with task success through safe imitation learning (IL) from demonstrations collected while safely performing multi-task. Fig. \ref{fig6a:urban_driving_results} shows the RR measured according to the number of expert demonstrations used for training, as well as the sum of the four CVs presented in Fig. \ref{fig4b:driving_constraints}. Our proposed DIAL method achieves both higher task performance and lower CV as more demonstrations are used, demonstrating the best results among the compared baselines. While there is a slight increase in the variance of RR in the last segment of the plot, the values show a gradually stable increase. This can be interpreted as a natural phenomenon due to the differences in scale among the various tasks. COCL maintains relatively stable performance but is not as efficient as DIAL. MERL shows little improvement in RR even when many demonstrations are used, and it decreases at the end due to overfitting. Additionally, CV remains high. Although MECL effectively reduces CV, RR decreases as more demonstrations are added due to excessive conservatism. 

To confirm whether DIAL has successfully learned all the designed constraints when trained with 300 expert trajectories, Fig. \ref{fig7:multi_constraints_learning_curves} presents learning curves showing the violation rates for each condition. As training progresses, the violation rates for all constraints gradually decrease and ultimately stabilize at low levels, demonstrating the stability and reliability of the proposed method. Furthermore, to examine the proposed method that approximates each constraint budget $\hat{\epsilon}_{i}$ as variables of a Beta distribution, Fig. \ref{fig8:constraint_distribution} shows how the density function of this distribution evolves for training. These results indicate that DIAL can learn to adhere to the constraints more effectively. At the beginning of training, all conditions start with low parameter values, resulting in a distribution that is evenly spread across the entire threshold range. As training progresses and the $\alpha_{2}$ value increases rapidly, the distribution becomes concentrated in the lower threshold region. This rapid increase can cause the agent to become overly conservative, potentially limiting task performance. To prevent this, the proposed method distorts the distribution based on randomly sampled risk levels, encouraging flexible exploration for specific conditions. Additionally, the term in Eq. \ref{eq8:constraint_regularizer} prevents overfitting of the distribution by incorporating a given prior probability. Due to this design, soft constraints like speeding or maintaining distance from the vehicle ahead can allow some violations during the middle stages of training, enabling finer adjustments. Towards the later stages of training, there is a noticeable tendency for the density to increase at specific thresholds and for the distribution to narrow. This implies that the model has gradually achieved stable performance.

\begin{table*}[!ht]
    \centering
    \caption{Safe IL Results on Robot Control Tasks}
    \resizebox{\textwidth}{!}{
        \begin{tabular}{c  c c c c c c  c c c c c c  c c c c c c}
            \toprule
            Environments 
            & \multicolumn{6}{c}{MountainCar}
            & \multicolumn{6}{c}{CartPole}
            & \multicolumn{6}{c}{BasicNav} \\
            \cmidrule(lr){2-7} \cmidrule(lr){8-13} \cmidrule(lr){14-19}
            
            $\#$ Trajectories
            & \multicolumn{2}{c}{1} 
            & \multicolumn{2}{c}{10} 
            & \multicolumn{2}{c}{50}
            
            & \multicolumn{2}{c}{1} 
            & \multicolumn{2}{c}{10} 
            & \multicolumn{2}{c}{50}
            
            & \multicolumn{2}{c}{1} 
            & \multicolumn{2}{c}{10} 
            & \multicolumn{2}{c}{50} \\
            
            \cmidrule(lr){2-3} \cmidrule(lr){4-5} \cmidrule(lr){6-7} \cmidrule(lr){8-9} \cmidrule(lr){10-11} \cmidrule(lr){12-13} \cmidrule(lr){14-15} \cmidrule(lr){16-17} \cmidrule(lr){18-19} 
            
            Metrics
            & SE $\uparrow$ & CR $\downarrow$
            & SE $\uparrow$ & CR $\downarrow$
            & SE $\uparrow$ & CR $\downarrow$
    
            & SE $\uparrow$ & CR $\downarrow$
            & SE $\uparrow$ & CR $\downarrow$
            & SE $\uparrow$ & CR $\downarrow$
    
            & SE $\uparrow$ & CR $\downarrow$
            & SE $\uparrow$ & CR $\downarrow$
            & SE $\uparrow$ & CR $\downarrow$  \\              
            
            \midrule
            MERL 
            & \textbf{4.17} & 4.03 
            & 4.21 & 0.34 
            & 4.30 & 0.18 
    
            & 4.41 & 7.25 
            & \underline{4.48} & 1.97 
            & \underline{4.49} & 1.41 
            
            & 2.61 & 85.3 
            & \textbf{2.89} & 47.2 
            & \textbf{2.97} & 17.9 \\ 
    
            MECL 
            & 3.45 & 0.03 
            & 4.08 & 0.12 
            & 4.10 & 0.61 
    
            & 4.35 & 1.21 
            & 4.46 & 1.81 
            & 4.47 & 2.02 
            
            & 1.61 & 0.38 
            & 1.96 & 1.01 
            & 2.02 & 1.75 \\ 
            
            DIAL
            & \underline{4.04} & \textbf{0.01} 
            & \textbf{4.29} & \textbf{0.02} 
            & \textbf{4.34} & \textbf{0.05} 
    
            & \textbf{4.45} & \textbf{0.03} 
            & \underline{4.48} & \textbf{0.26} 
            & \underline{4.49} & \textbf{1.37} 
            
            & \textbf{2.73} & \textbf{0.35} 
            & \underline{2.84} & \textbf{0.71} 
            & \underline{2.94} & \textbf{1.10} \\ 
            
            \bottomrule
        \end{tabular}
    }
    \label{table1:safeIL_results_on_robot_control}
\end{table*}

\begin{figure*}[!ht]
    \centering
    \subfloat[State Visitation Map]{
        \includegraphics[width=0.4\textwidth]{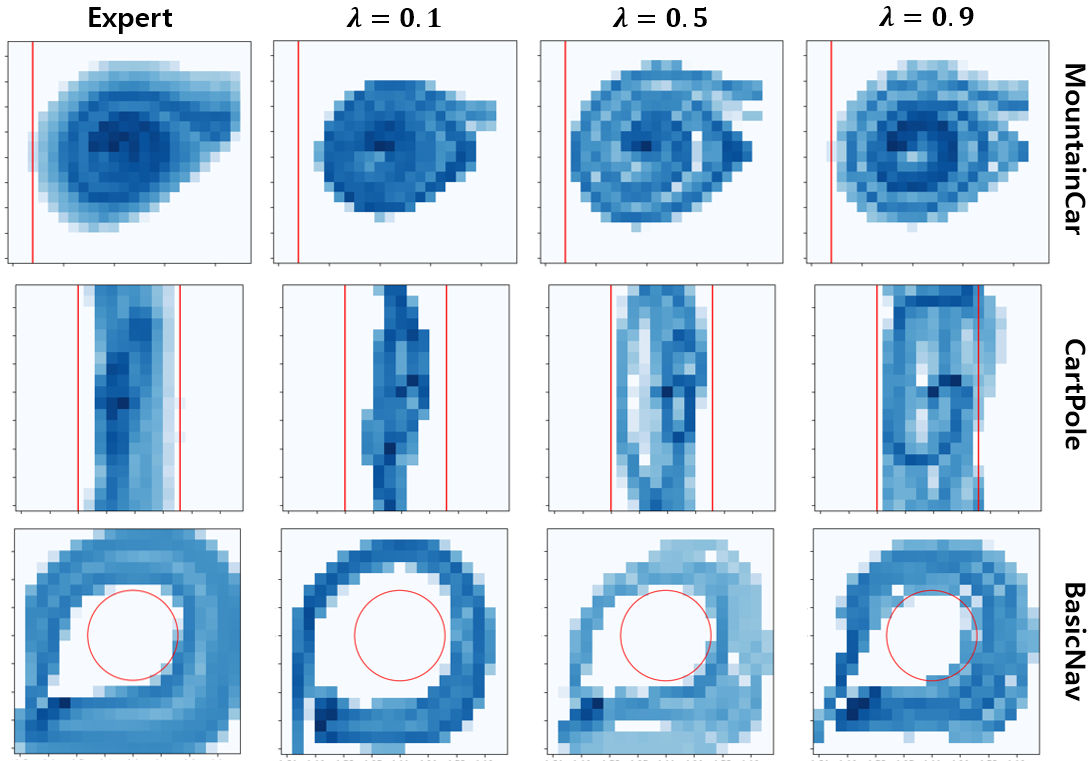}
        \label{fig9a:visitation_map_safeIL}
    } 
    \hfill
    \subfloat[Inferred Constraint Map]{
        \includegraphics[width=0.57\textwidth]{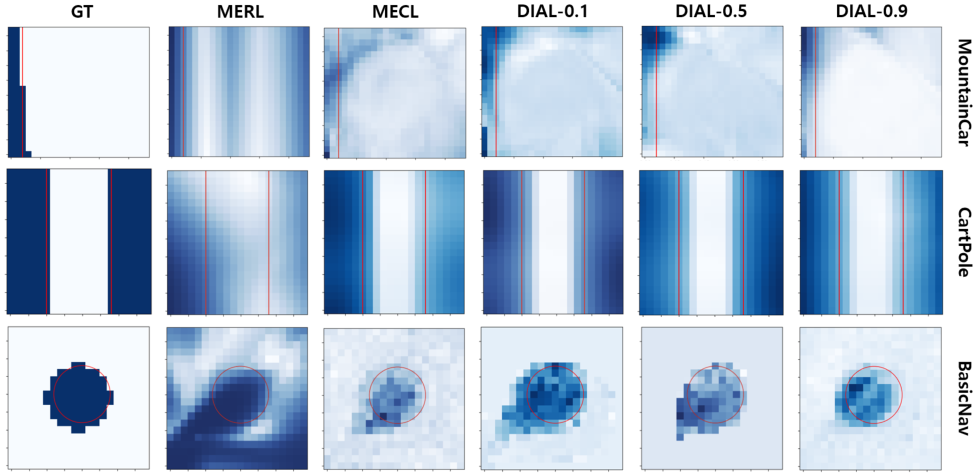}
        \label{fig9b:constraint_map_safeIL}
    }
    \caption{Visualization of state visitation frequencies and inferred constraints. Each map is normalized between 0 and 1, and darker blue indicates higher values.}
    \label{fig9:visualization_map_safeIL}
\end{figure*}

In the robot control environments, we evaluate the degree of safe exploration of policy by comparing the SE and CR based on the number of expert trajectories used, as shown in Tab. \ref{table1:safeIL_results_on_robot_control}. DIAL demonstrates higher SE and simultaneously lower CR compared to other methods, even when using a relatively small number of trajectories in all environments. MERL exhibits high exploration performance in terms of SE but has a high CR, leading to reduced safety. MECL maintains a relatively low CR but has low SE, indicating insufficient exploration performance. Although DIAL's SE is slightly lower than MERL's, the difference is minimal as shown by the underlined numbers and there is a significant difference in CR. These results demonstrate that our method successfully balances diverse state exploration and safety.

\begin{figure}[!t]
    \centering
    \includegraphics[width=1.0\linewidth]{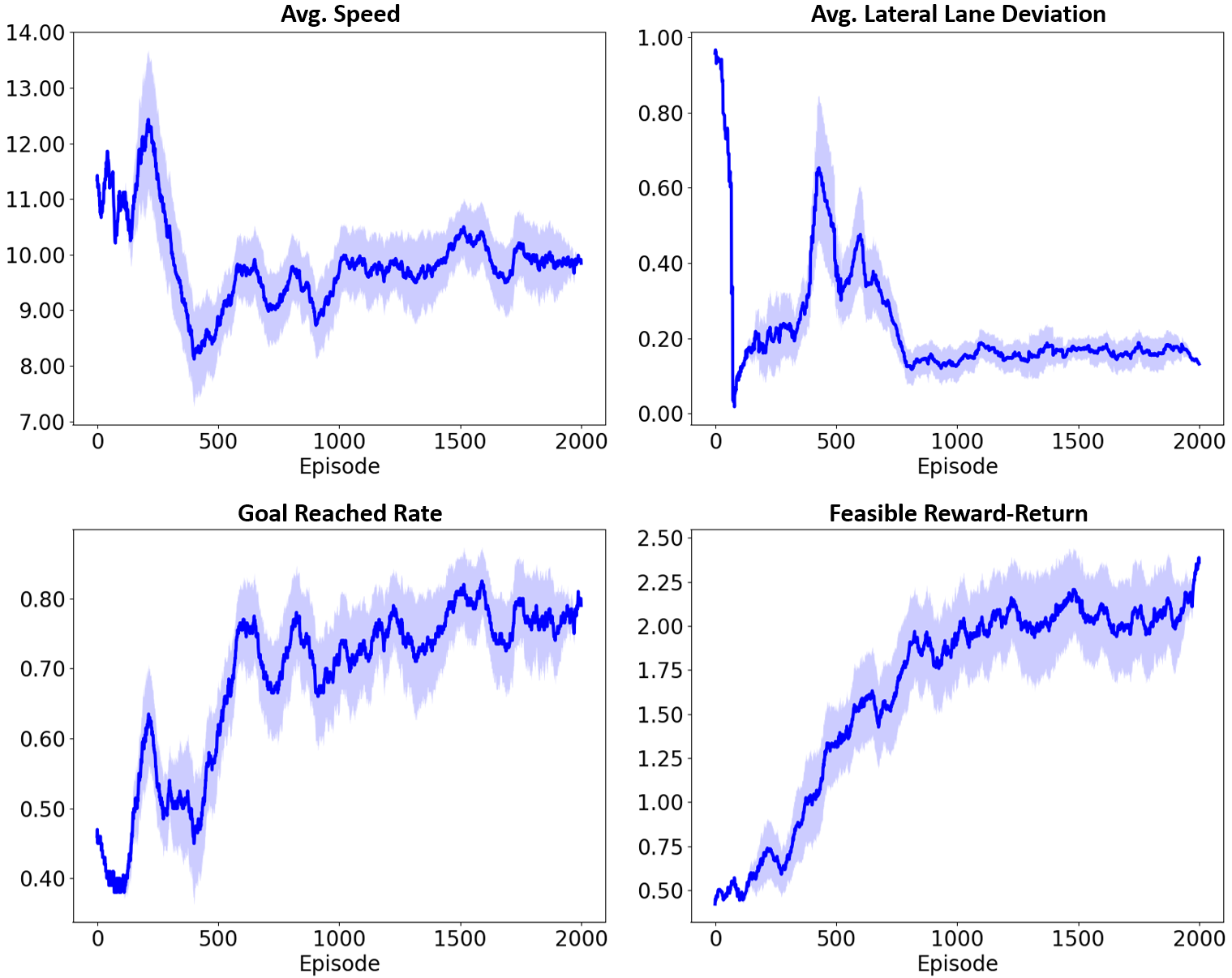}  
    \caption{Learning curves of DIAL showing environmental data on safety and task success during safe TL.}
    \label{fig:meta_learning_curves}
\end{figure}

Furthermore, Fig.~\ref{fig9:visualization_map_safeIL} visually demonstrates the effectiveness of DIAL's design in finely adjusting the deviation in risk levels by carefully selecting $\lambda$ with limited data. Even when using the same expert data shown on the far left in Fig.~\ref{fig9a:visitation_map_safeIL}, the exploratory tendencies of the learned policy vary depending on the choice of $\lambda$. Policy tend to avoid risks when $\lambda$ is low and take risks when $\lambda$ is high. This result suggests that setting a low $\lambda$ is advantageous for balancing when expert states are near the safety boundary. Conversely, when the data is far from the boundary, setting a high $\lambda$ is more appropriate. Fig.~\ref{fig9b:constraint_map_safeIL} shows that we can infer a constraint that most closely resembles the ground truth (GT) located on the far left by finely tuning $\lambda$ in DIAL. In contrast, MERL fails to properly capture the constraint distribution, and MECL can have its distribution's center and shape distorted differently from the GT due to reliance on the characteristics of the data used for training.

\begin{figure*}[!ht]
    \centering
    \includegraphics[width=1.0\linewidth]{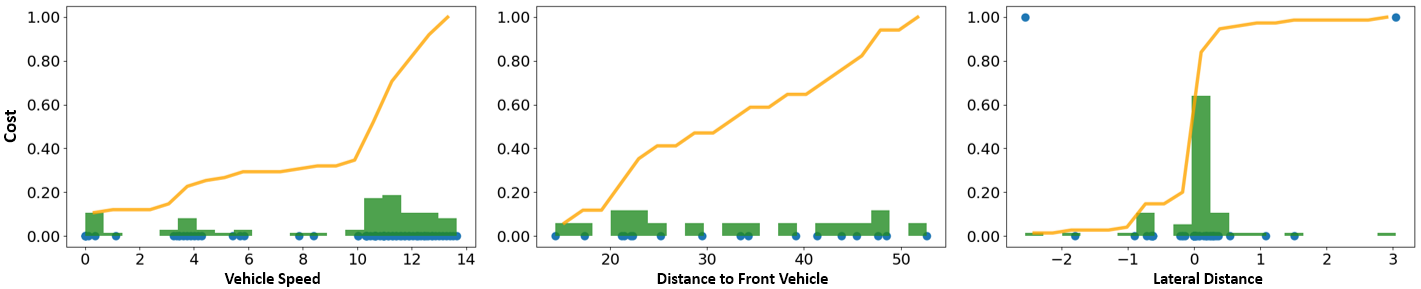}  
    \caption{Visualization of the distribution of environmental data related to safety requirements during a single episode executed by an agent trained with DIAL. The blue dots represent the cost values corresponding to the x-axis data, which are used to evaluate the constraints. The green bars indicate the histogram of the x-axis data. The yellow line represents the CDF corresponding to the histogram.}
    \label{fig:episode_information}
\end{figure*}

\begin{table*}[!t]
    \centering
    \caption{Safe TL Results on Robot Control Tasks}
        \begin{tabular}{c  c c  c c  c c  c c}
            \toprule
            Environments
            & \multicolumn{2}{c}{MountainCar}
            & \multicolumn{2}{c}{CartPole} 
            & \multicolumn{2}{c}{BasicNav} 
            & \multicolumn{2}{c}{PointGoal} \\
            \cmidrule(lr){2-3} \cmidrule(lr){4-5} \cmidrule(lr){6-7} \cmidrule(lr){8-9}
            Metrics
            & RR $\uparrow$ & CR (0.5) $\downarrow$  
            & RR $\uparrow$ & CR (5) $\downarrow$
            & RR $\uparrow$ & CR (10) $\downarrow$
            & RR $\uparrow$ & CR (25) $\downarrow$ \\
            \midrule
            MERL
            & 74.2 $\pm$ 12.6
            & 4.26 $\pm$ 1.63
            
            & \textbf{695 $\pm$ 1.34}
            & 99.4 $\pm$ 26.7

            & \textbf{215 $\pm$ 0.64}
            & 98.7 $\pm$ 0.17

            & \textbf{13.3 $\pm$ 0.09}
            & 34.3 $\pm$ 0.85 \\
            
            MECL
            & 36.2 $\pm$ 15.2
            & 0.64 $\pm$ 0.28

            & \underline{694 $\pm$ 1.45}
            & 86.1 $\pm$ 19.4

            & 210 $\pm$ 0.83
            & 74.7 $\pm$ 12.6

            & 7.08 $\pm$ 0.23
            & 27.5 $\pm$ 1.94 \\
            
            DIAL
            & \textbf{92.9 $\pm$ 0.23} 
            & \textbf{0.41 $\pm$ 0.11} 

            & \underline{693 $\pm$ 2.25}
            & \textbf{3.66 $\pm$ 2.01}

            & \underline{213 $\pm$ 0.54}
            & \textbf{4.32 $\pm$ 2.88}

            & \underline{10.55 $\pm$ 0.39}
            & \textbf{20.3 $\pm$ 1.19} \\
            \bottomrule
        \end{tabular}    
    \label{table2:safeTL_results_on_robot_control}
\end{table*}

\subsection{Results for Safe Transfer Learning}

In this section, we address safe TL in environments where the safety requirements are the same as safe IL, but the explicit reward functions for the target task are given. In safe TL, the key concern is whether the agent can maintain compliance with the constraints without forgetting them despite changes in the objective function for solving the target task. To demonstrate that using the function recovered through safe IL reduces the burden of cost design, we use the extrinsic cost only to evaluate safety and do not use it in safe TL. Details of the hyperparameters used for training the agent are provided in Tab. \ref{table:safeTL_hyperparameters} in Appendix. \ref{appendixB:implementation_details}. Fig. \ref{fig6b:urban_driving_results} compares the performance of meta-task learning in the urban driving environment illustrated in Fig. \ref{fig4c:meta-task_driving}. For DIAL, as the number of demonstrations increases, RR steadily improves while CV decreases. In the case of COCL, although it has slightly better performance than DIAL when the number of demonstrations is small, it shows only a marginal reduction in CV and almost no improvement in RR, even with an increased number of demonstrations. MERL has significant limitations in terms of safety, while MECL suffers from degraded performance due to overly conservative behavior. Overall, DIAL effectively balances performance and safety, utilizing the available demonstrations to achieve the best overall results.

\begin{figure}[!t]
    \centering
    \includegraphics[width=1.0\linewidth]{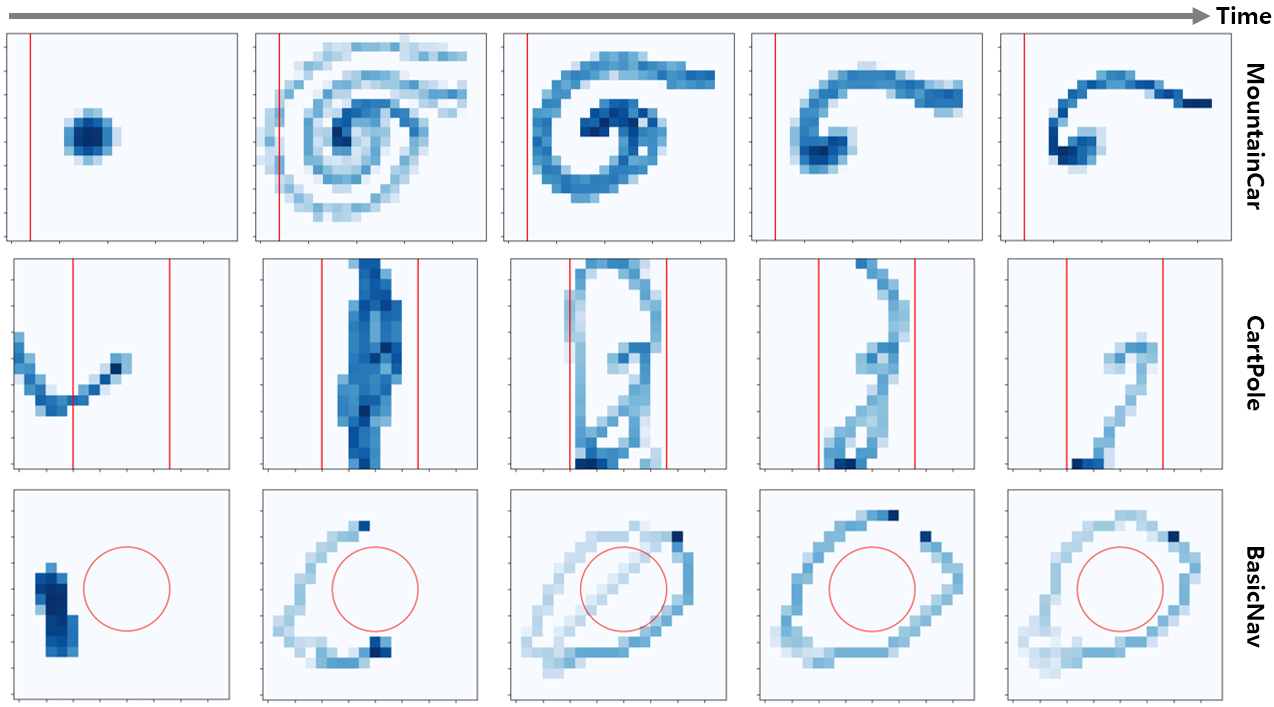}  
    \caption{Visualization of changes in state visitation maps during safe TL}
    \label{fig:transfer_target}
\end{figure}

Fig. \ref{fig:meta_learning_curves} illustrates how the DIAL agent gradually acquires the ability to perform tasks safely and efficiently during the learning process. The average speed fluctuates significantly in the early stages, However, it gradually stabilizes between approximately 11 and 12 as the episodes progress. The vehicle's deviation from the lane center also reaches a stable level below 0.2. The goal achievement rate and feasible rewards steadily increase and converge to certain values, where feasible rewards are the sum of rewards the agent obtains only when all safety requirements are met. In the $20$\% of cases where the goal is not achieved, timeouts occur due to deadlocks caused by congestion or collisions in the surrounding traffic. To verify how the agent trained with DIAL performs the meta-task safely, we visualize the distribution of environmental data obtained by the agent while executing tasks in a single episode, as shown in Fig. \ref{fig:episode_information}. The vehicle's speed remains mostly concentrated near the threshold without exceeding the speed limit of 15. The distance to the vehicle ahead is evenly distributed over values greater than the threshold of 10. The lateral distance from the center of the lane is concentrated within the range of -1 to 1 to prevent lane departure, where the deviations in values are due to reaching the destination. These results show that the agent learns and retains behaviors such as maintaining speed limits, safe distances, and staying centered in the lane, even when performing novel tasks.

Tab. \ref{table2:safeTL_results_on_robot_control} presents a comparison of RR and CR measured using the fully trained policies on robot control tasks. DIAL consistently exhibits the lowest CR across all environments, satisfying the condition that the average is less than the threshold for each environment, as indicated in parentheses. In terms of RR, DIAL shows the highest performance in MountainCar, and while MERL has the highest RR in the other environments, DIAL achieves performance that is nearly comparable to MERL, as seen in the underlined numbers. For MERL, although the RR is relatively high, the CR is extremely large, which limits its ability to address safety issues. In the case of MECL, the RR is generally low, and the CR is higher than that of DIAL, exceeding the threshold for each environment. 

Fig. \ref{fig:transfer_target} visually depicts the improvement process of the policy learned through DIAL in safe TL. In the early stages of training, the agent visits as many safe states as possible. During the middle of training, unsafe interactions occur at the boundaries of the safe area. However, The agent gradually converges to maximize the reward for the given target task while satisfying safety constraints. For detailed information on the rewards for each environment, please refer to Figure \ref{fig:explcit_function} in Appendix \ref{appendixA:environmental_settings}. These results demonstrate that even when the objective function is altered to maximize the reward of a new target task, the proposed method does not lose the safety information previously learned.

\section{Conclusion}
This paper proposes a novel approach called DIAL for safe RL in autonomous driving. DIAL leverages multi-task demonstrations to reconstruct the distribution of shared safety constraints and flexibly adjusts the required risk levels to address new tasks, demonstrating superior safety and efficiency compared to existing methods in experimental results. This approach offers a promising solution for safe exploration in safety-critical autonomous systems by enabling safe adaptation to new environments without relying on explicitly defined constraints. However, DIAL requires sufficient demonstrations to learn scalable constraints across multiple tasks, which can be challenging and costly in complex environments. Additionally, the two-stage learning structure, in which constraints are first learned from data and then used to safely adapt based on the given reward functions, can reduce learning efficiency. Moreover, selecting inappropriate risk levels during the distortion of constraint distributions may lead to overly conservative or overly optimistic behaviors, potentially hindering task performance or compromising safety. To overcome these limitations, it is necessary to develop methods that effectively utilize suboptimal demonstrations. Furthermore, integrating the two-stage learning structure into a single stage can enhance learning efficiency. Additionally, incorporating techniques that automatically optimize or dynamically adjust risk levels could achieve a more effective balance between safety and performance, presenting a promising research direction. Extending these approaches to other safety-critical domains, such as healthcare, would also allow for the validation of their scalability and broad applicability. These research directions are expected to overcome the limitations of DIAL and contribute to the development of more robust and efficient safe RL methods.

\appendices

\begin{table*}[!ht]           
    \caption{Environmental Configurations}
    \centering
    \begin{adjustbox}{max width=1.0\textwidth}
        \begin{tabular}{c c c c c c}
            \toprule
            Configurations
            & Intersection
            & MountainCar 
            & CartPole 
            & BasicNav 
            & PointGoal \\
            \midrule
            State Dimension 
            & [15, 7]
            & 2
            & 4
            & 2 
            & 36 \\                        
            Action Dimension 
            & 5
            & 1
            & 1
            & 2
            & 2 \\
            Constraint Budget ($\epsilon$)
            & [0.2, 0.2, 0.05, 0.1]
            & 0.005
            & 0.05
            & 0.1
            & 0.25 \\
            Maximum Episode Length ($T$)
            & 75
            & 400
            & 400
            & 1200
            & 500 \\
            Discretized States 
            & - 
            & [Position, Velocity] 
            & [Cart Position, Pole Angle]
            & [Position X, Position Y] 
            & - \\
            Size for Discretization ($M$)
            & - 
            & [24, 22]
            & [20, 20]
            & [20, 20]
            & - \\
            Reward Function
            & $10 \times \mathbb{1}_{\text{Goal}} + \xi_{v} v_{t} + \xi_{\angle}\vert \angle_{t}\vert$
            & $100 \times \mathbb{1}_{\{s_{t}=s_{goal}\}} - 0.1a_{t}^{2}$
            & $1 + cos\; \theta_{t}$
            & $100 \times (d_{t-1} - d_{t})$
            & $(d_{t-1} - d_{t})$ \\
            \bottomrule
            \label{table:environmental_settings}
        \end{tabular}
    \end{adjustbox}            
\end{table*} 

\begin{figure*}[!ht]    
    \vspace{-0.5cm}
    \centering
    \subfloat[MountainCar]{
        \centering
        \includegraphics[width=0.22\textwidth]{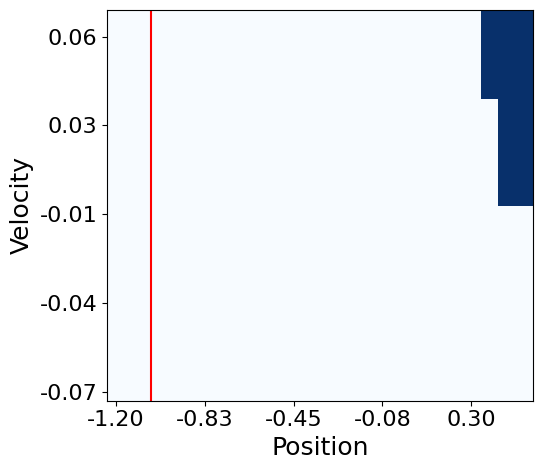}
    } 
    \subfloat[CartPole]{
        \centering
        \includegraphics[width=0.21\textwidth]{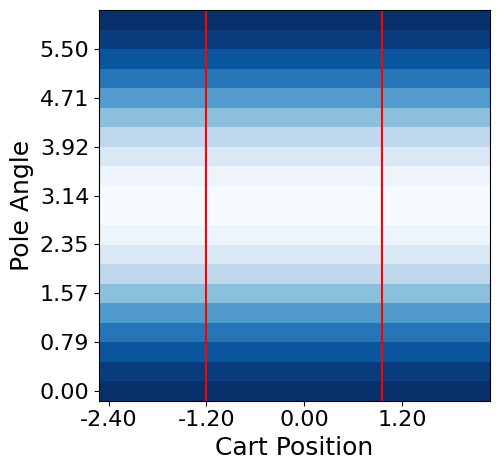}
    }
    \subfloat[BasicNav]{
        \centering
        \includegraphics[width=0.21\textwidth]{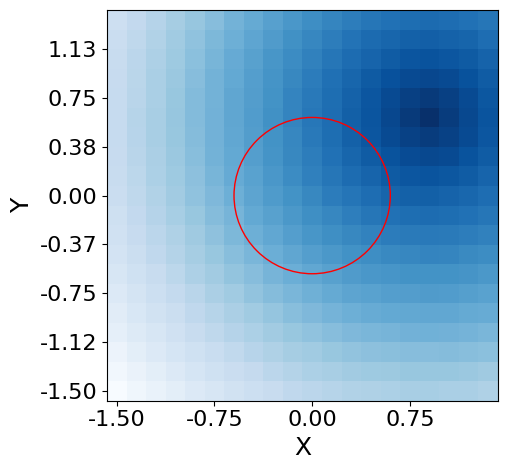}
    }       
    \caption{
        Visualization of explicit reward for each environment. Darker blue represents higher values for each map.
    }
    \label{fig:explcit_function}
\end{figure*}

\section{Environmental Settings}
\label{appendixA:environmental_settings}
Tab. \ref{table:environmental_settings} provides details about each environment used in the experiments. Although we assume infinite-horizon settings, the experimental environment is finite-horizon, which requires calculating the discounted approximation of \( d \) in a finite horizon \( T \) as \( \epsilon = \frac{(1 - \gamma)d}{1 - \gamma^{T}} \). The parameter \( d \) represents the threshold for the cost-return. The value \( \epsilon \) ranges between 0 and 1. It indicates the probability that the agent violates the constraint in a single episode. For the robot control tasks shown in Fig. \ref{fig5:robot_control}(a-d), \(d\) is set to 0.5, 5, 10, and 25, respectively. These values are consistent with the configurations in \cite{yang2023cem}. In high-dimensional environments such as Intersection and PointGoal, the states are not discretized for visualization because selecting two main dimensions that clearly represent the state space is difficult. To aid understanding, Fig. \ref{fig:explcit_function} presents the visualization map of the reward function based on discretized states for each environment depicted in Fig. \ref{fig5:robot_control}(a-c). In the main text, the axis information for these visualization maps is consistent with that in the figure and is omitted for simplicity. To implement the PointGoal environment using the SafetyGym engine, the configuration dictionary is as follows:

\begin{lstlisting}[
frame=lines,                    % 코드 블록 위아래에 선 추가
backgroundcolor=\color{LightGray}, % 배경색 설정
basicstyle=\footnotesize\ttfamily, % 글꼴 크기 및 타입 설정
numbers=left,                   % 줄 번호를 왼쪽에 표시
numberstyle=\tiny\color{gray},  % 줄 번호 스타일 설정
xleftmargin=1em,                % 왼쪽 여백
xrightmargin=1em,               % 오른쪽 여백
language=Python,                % 기본 프로그래밍 언어 설정
breaklines=true,                % 자동 줄 바꿈
showstringspaces=false          % 문자열의 공백 표시 안함
]
    import safety_gym
    from gym.envs.registration import register
    
    register(id='PG-v0',
         entry_point='safety_gym.envs.mujoco:Engine',
         max_episode_steps=500,
         kwargs={'config': pointgoal_config})
         
    pointgoal_config = {
        'task': 'goal',
        'robot_base': 'xmls/point.xml',
        'observe_goal_lidar': True,
        'observe_box_lidar': True,
        'lidar_max_dist': 3,
        'lidar_num_bins': 8,
        'goal_size': 0.3,
        'goal_keepout': 0.305,
        'hazards_size': 0.2,
        'hazards_keepout': 0.1,
        'constrain_hazards': True,
        'observe_hazards': True,
        'observe_vases': True,
        'placements_extents': [-1.5, -1.5, 1.5, 1.5],
        'hazards_num': 8,
        'vases_num': 1}
\end{lstlisting}

\section{Implementation Details}
\label{appendixB:implementation_details}

\textbf{Expert Demonstrations:}
To collect the expert trajectories $\mathcal{D}_{E}$ for urban driving tasks, the agent is trained to maximize true rewards while satisfying multiple constraints defined by the designed features in Fig. \ref{fig4b:driving_constraints} and the ground-truth budgets. The trained agent from \cite{wen2018constrained} is then deployed in the environment shown in Fig. \ref{fig4a:multi-task_driving}, targeting randomly selected goal positions that require either a left or right turn. For robot control tasks, we deploy a trained agent based on \cite{yang2023cem} in an environment with known true costs but no assigned target task.

\textbf{Hyperparameters:}
This section briefly describes the hyperparameters required to reproduce our experiments. As shown in Tab. \ref{table:unified_hyperparameters}, several hyperparameters are unified across all methods to ensure a fair comparison. The parameters \(n_{samp}\), \(n_{elite}\), and \(n_{iter}\) correspond to the hyperparameters of CEM. While these are not explicitly mentioned in the main text, the pseudocode for CEM is provided in Algo. 4 of \cite{lindner2024learning}. Tab. \ref{table:safeIL_hyperparameters} and \ref{table:safeTL_hyperparameters} present the hyperparameters used for each environment during the safe IL and safe TL stages, respectively.

\begin{table}[!ht]                    
    \caption{Hyperparameters Unified in ALL Experiments}
    \centering
        \begin{tabular}{c c c}
            \toprule
            Hyperparameters
            & Value
            & Notation \\
            \midrule
            Number of $\tau$ in Rollout Buffer $\mathcal{D}$
            & $20$
            & $N$ \\            
            Discount Factor
            & 0.99
            & $\gamma$ \\
            Learning Rate for Constraint
            & $1 \times 10^{-2}$
            & $\eta_{C}$ \\
            Learning Rate for Constraint Prior
            & $1 \times 10^{-2}$
            & $\eta_{P}$ \\
            Learning Rate for Safety Weight
            & $1 \times 10^{-3}$
            & $\eta_{\kappa}$ \\
            Learning Rate for Reward
            & $1 \times 10^{-3}$
            & $\eta_{R}$ \\            
            Initial Safety Weight            
            & 1.0
            & $\kappa_{0}$ \\
            Number of Neighbors
            & 4
            & $k$ \\
            Damp Scaling Factor            
            & 10
            & $\kappa_{d}$ \\
            Beta Prior
            & [0.1, 0.9]
            & $\alpha_{0}$ \\
            Number of Samples for CEM
            & 80
            & $n_{samp}$ \\
            Number of Elites for CEM
            & 20
            & $n_{elite}$ \\
            Number of Iterations for CEM
            & 5
            & $n_{iter}$ \\
            \bottomrule
            \label{table:unified_hyperparameters}
        \end{tabular}
\end{table}  

\begin{table*}[!ht]                    
    \caption{Hyperparameters for Each Environment in Safe IL}
    \centering
        \begin{tabular}{c c c c c c c}
            \toprule
            Hyperparameters
            & Intersection
            & MountainCar 
            & CartPole 
            & BasicNav 
            & PointGoal 
            & Notation \\
            \midrule
            Environmental Steps 
            & $1.5 \times 10^{5}$
            & $5 \times 10^{4}$
            & $5 \times 10^{5}$
            & $2 \times 10^{5}$
            & $10^{6}$
            & - \\
            Entropy Coefficient
            & 0.01
            & 0.01
            & 0.01
            & 1.0
            & 0.1
            & $\beta$ \\
            Trust-Region Threshold
            & $0.1$
            & $0.5$
            & $0.5$
            & $1.0$
            & $0.1$
            & $\delta$ \\
            \bottomrule
            \label{table:safeIL_hyperparameters}
        \end{tabular}
\end{table*}  

\begin{table*}[!ht]                    
    \caption{Hyperparameters for Each Environment in Safe TL}
    \centering
    \begin{adjustbox}{max width=1.0\linewidth}
        \begin{tabular}{c c c c c c c}
            \toprule
            Hyperparameters
            & Intersection            
            & MountainCar 
            & CartPole 
            & BasicNav 
            & PointGoal 
            & Notation \\
            \midrule
            Environmental Steps 
            & $1.5 \times 10^{5}$
            & $5 \times 10^{4}$
            & $5 \times 10^{5}$
            & $5 \times 10^{5}$
            & $1.5 \times 10^{6}$
            & - \\          
            Number of $\tau_{E}$ in Expert Buffer $\mathcal{D}_{E}$
            & 100
            & 50
            & 50
            & 50
            & 500
            & $N_{E}$ \\
            Risk Level
            & 0.5
            & 0.5
            & 0.5
            & 0.5
            & 0.1
            & $\lambda$ \\
            Entropy Coefficient
            & \multicolumn{5}{c}{0.01}
            & $\beta$ \\
            \bottomrule
            \label{table:safeTL_hyperparameters}
        \end{tabular}
    \end{adjustbox}            
\end{table*}  

\begin{figure*}[!ht]
    \centering
    \begin{adjustbox}{max width=1.0\textwidth}
        \begin{tabular}{c c c}
             \includegraphics[width=0.25\linewidth]{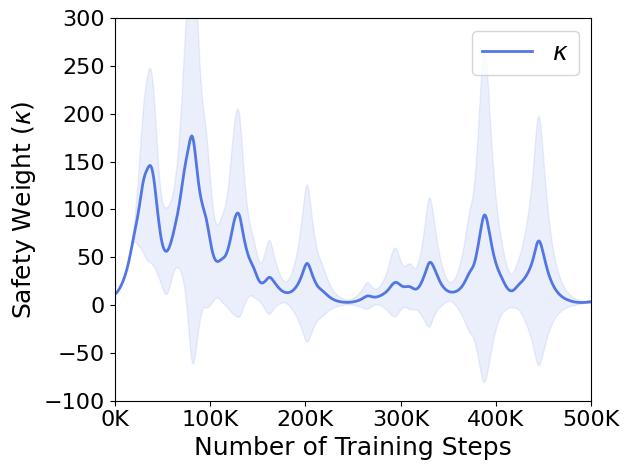} \hspace{-0.5cm} &
             \includegraphics[width=0.25\linewidth]{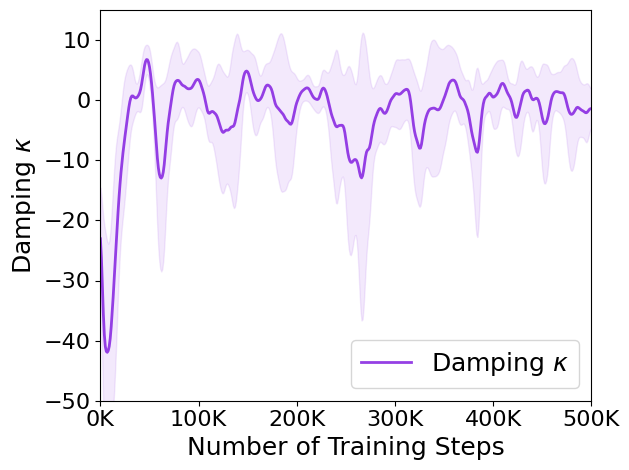} \hspace{-0.5cm} &
             \includegraphics[width=0.25\linewidth]{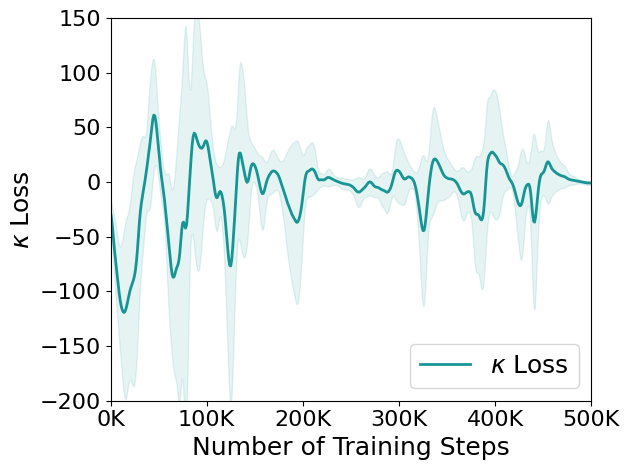}
        \end{tabular}
    \end{adjustbox}
    \caption{Learning curves for auto-tuning safety weight.}
    \label{fig:safety_weight}
\end{figure*}

\textbf{Technique for Learning Stability:} In general, the learning rate of $\kappa$ is set to a small value. At this point, the following issues arise. When the policy is unsafe, $\kappa$ cannot quickly adjust to a larger value needed to ensure safety. Conversely, when the policy is safe, $\kappa$ does not swiftly revert to a smaller value. This fact destabilizes the learning process. To address the instability, we introduce a damping weight $\tilde{\kappa}$ in place of $\kappa$ in Eq. \ref{eq11:policy_objective}. This adjustment directly tackles the issue by allowing the algorithm to dynamically respond to safety considerations while stabilizing the learning process. The damping weight is defined as $\tilde{\kappa} = \kappa - \kappa_{d}  \big(\epsilon - \mathbb{E}_{\tau \sim \pi_{\theta}(\cdot)}[\bar{\Gamma}_{\phi}^{\lambda}(\tau)])$ based on methods \cite{platt1987constrained, kumar2021controlled}, where $\kappa_{d}$ is a damp scaling factor.  This formulation adjusts 
$\kappa$ based on the difference between the safety threshold $\epsilon$ and the expected safety risk $\mathbb{E}_{\tau \sim \pi_{\theta}(\cdot)}[\bar{\Gamma}_{\phi}^{\lambda}(\tau)]$, providing a more nuanced response to environmental conditions. Fig. \ref{fig:safety_weight} illustrates how the original safety weight $\kappa$ and the damping weight $\tilde{\kappa}$ operate during environmental interactions, as measured in the CartPole environment. The loss value, which becomes negative when the expected risk exceeds the safety limit and positive when it falls below the limit, directly influences these weights. When the loss value is negative, the safety weight $\kappa$ increases, but the damping weight $\tilde{\kappa}$ prevents a sharp rise. Conversely, when the loss value is positive, the safety weight decreases, and the damping weight mitigates a rapid decline. Over time, the loss value and damping weight stabilize around zero, while the safety weight fluctuates before eventually converging to a stable value. This approach ensures a more stable learning process by dynamically adjusting $\tilde{\kappa}$ based on safety requirements at each iteration.

\bibliographystyle{IEEEtran}
\bibliography{IEEEabrv, ref}

\vspace{11pt}

\begin{IEEEbiography}[{\includegraphics[width=1in,height=1.25in,clip,keepaspectratio]{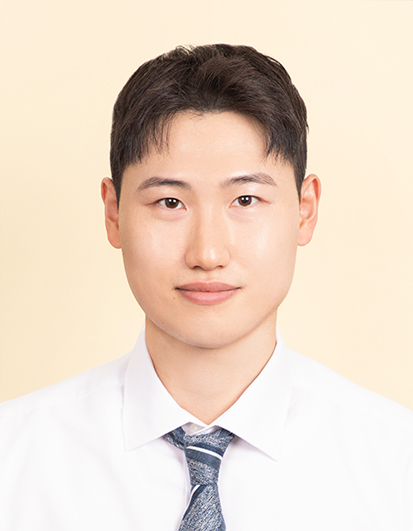}}]{Se-Wook Yoo}
(Member, IEEE) received the B.S. degree in electrical and electronic engineering from Hongik University, Seoul, South Korea, in 2018. He is currently pursuing the Ph.D. degree in electrical and computer engineering from Seoul National University, Seoul, South Korea. His research interests include reinforcement learning, imitation learning, and autonomous driving.
\end{IEEEbiography}

\vspace{11pt}

\begin{IEEEbiography}[{\includegraphics[width=1in,height=1.25in,clip,keepaspectratio]{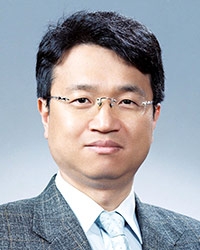}}]{Seung-Woo Seo}
(Member, IEEE) received the B.S. and M.S. degrees in electrical engineering from Seoul National University, Seoul, South Korea, and
the Ph.D. degree in electrical engineering from The Pennsylvania State University, University Park, PA, USA. In 1996, he joined as a Faculty Member with the School of Electrical Engineering, Institute of New Media and Communications and the Automation and Systems Research Institute, Seoul National University. He was a Faculty Member at the Department of Computer Science and Engineering, The Pennsylvania State University. He also worked as a member of the Research Staff at the Department of Electrical Engineering, Princeton University, Princeton, NJ, USA. He is currently working as a Professor of electrical engineering with Seoul National University and the Director of the Intelligent Vehicle IT (IVIT) Research Center funded by Korean Government and Automotive Industries.
\end{IEEEbiography}

\vfill

\end{document}